\documentclass[a4paper,12pt]{article}
\usepackage[fleqn]{amsmath}
\usepackage[english]{babel}
\usepackage{amssymb}
\usepackage{harvard}
\usepackage{graphicx}
\usepackage{epsfig}      
\usepackage{graphics}
\usepackage{overpic}
\usepackage{rotating}
\usepackage{multirow}
\usepackage{subcaption}
\usepackage{bbm}
\usepackage{url}
\usepackage{hyperref}
\usepackage{tikz}
\usepackage{comment}
\usepackage{xcolor}
\usepackage[export]{adjustbox}
\usepackage{algorithm}
\usepackage{algpseudocode}
\usepackage{framed}
\usepackage{makecell}
\usepackage{booktabs}
\numberwithin{equation}{section}
\usepackage[titletoc,toc,title,page]{appendix}
\usepackage[flushleft]{threeparttable}

\def\bee{\begin{eqnarray}}
\def\eee{\end{eqnarray}}
\def\beee{\begin{eqnarray*}}
\def\eeee{\end{eqnarray*}}

\def\be{\begin{equation}}
\def\ee{\end{equation}}

\begin{document}

  \citationstyle{dcu}
  \bibliographystyle{dcu}
  \citationmode{default}

\setcounter{page}{0} \thispagestyle{empty}
\begin{center}
\begin{Large}
\bf Choosing the Number of Topics in LDA Models -- A Monte Carlo Comparison of Selection Criteria\footnote{Financial support from the German Research Foundation (DFG) (WI 2024/8-1) and the National Science Centre (NCN) (Beethoven Classic 3: UMO-2018/31/G/HS4/00869) for the project TEXTMOD is gratefully acknowledged. The project also benefited from cooperation within HiTEC Cost Action CA 21163. }\\[30pt]
\end{Large}

\begin{small}
{Victor Bystrov}\\
University of Lodz\\
Rewolucji 1905r. 41, 90-214 Lodz, Poland\\
email: victor.bystrov@uni.lodz.pl\\[2ex]

{Viktoriia Naboka-Krell}\\
Justus Liebig University Giessen\\
Licher Strasse 64, 35394 Giessen, Germany\\
email: viktoriia.naboka@wirtschaft.uni-giessen.de\\[2ex]

{Anna Staszewska-Bystrova}\\
University of Lodz\\
Rewolucji 1905r. 37/39, 90-214 Lodz, Poland\\
email: anna.bystrova@uni.lodz.pl\\[2ex]

{Peter Winker}\\
Justus Liebig University Giessen\\
Licher Strasse 64, 35394 Giessen, Germany\\
email: peter.winker@wirtschaft.uni-giessen.de
\end{small}

\end{center}

\noindent {\bf Abstract}

Selecting the number of topics in LDA models is considered to be a difficult task, for which alternative approaches have been proposed. The performance of the recently developed singular Bayesian information criterion (sBIC) is evaluated and compared to the performance of alternative model selection criteria. The sBIC is a generalization of the standard BIC that can be implemented to singular statistical models. The comparison is based on Monte Carlo simulations and carried out for several alternative settings, varying with respect to the number of topics, the number of documents and the size of documents in the corpora. Performance is measured using different criteria which take into account the correct number of topics, but also whether the relevant topics from the DGPs are identified. Practical recommendations for LDA model selection in applications are derived. \\


\newpage

\section{Introduction}\label{sec:intro}
Text data have been increasingly used in different applications lately. One of the main challenges in working with text data is to structure and to quantify these data. To this end, probabilistic topic modelling approaches are often applied, as they allow to uncover hidden structures behind text data. One of the best-known and widely used topic modelling approaches is Latent Dirichlet Allocation (LDA) introduced by~\citeasnoun{Blei2003}. For some recent applications making use of this method, see, e.g.,~\citeasnoun{Luedering2016}, \citeasnoun{Thorsrud2020}, \citeasnoun{Ellingsen2022}, and \citeasnoun{Savin2022}.

LDA is an unsupervised method that builds on two basic assumptions. First, it is assumed that each document in a corpus represents a mixture of topics. The second assumption is that each topic is given by a mixture of words from the vocabulary. The number of these topics, or themes, is a parameter to be set by the researcher. Often this decision is based on human/expert judgment and is, therefore, rather subjective. In order to account for possible subjectivity and to allow for a more standardised estimation procedure, different evaluation metrics have been developed for identifying an optimal number of topics in LDA models. Some of them aim to minimize the similarity of different topics~\cite{Cao2009}, maximize the topic coherence~\cite{Mimno2011} or maximize the goodness-of-fit between the estimated and the actual document-word frequencies~\cite{Grossetti2022}. These criteria, however, often result in (substantially) different numbers when applied to the same corpus. Their performance might also differ across corpora depending on the underlying data set (see examples in Section~\ref{sec:eval_metrics}). \citeasnoun{Bystrov2022} propose to use a new measure for selecting an optimal number of topics, namely the singular Bayesian information criterion (sBIC). This information criterion reflects the trade-off between goodness-of-fit and model complexity and showed promising results in a first application.

There have been some attempts to compare selected criteria based on individual real datasets.\footnote{A notable exception including also a small scale Monte Carlo simulation is \citeasnoun{Grossetti2022}.} In this paper, a comprehensive Monte Carlo (MC) simulation is proposed, which allows a systematic evaluation going beyond individual case reports by using a large number of datasets coming from well defined data generating processes (DGP) with known properties. Thereby, we consider three different data generating processes to reflect different types of text data commonly used in applications. In a first step, we generate corpora with a known (true) number of topics, to which LDA models with different numbers of topics are fitted. Then, we apply the metrics to select the number of topics and evaluate their performance over many MC replications. To the best of our knowledge, no such systematic and comprehensive comparison analysis of the metrics used for choosing the number of topics in LDA models has been performed yet.

The contribution of this paper is threefold. First, with the sBIC we implement a new measure for identifying the true number of topics in LDA models. Second, we perform a proper MC simulation study to evaluate the proposed criterion as well as other evaluation criteria commonly used in applications.\footnote{This version of the paper is preliminary regarding the limited number of replications conducted for the Monte Carlo simulations. Due to constraints in available computational resources, current work focuses on extending the number of replications to 1\,000 for all setups.} Third, we evaluate the considered metrics quantitatively and qualitatively, i.e., we consider whether the actual number of topics is approximated well, and also the content and structure of the estimated topics.

The remainder of this paper is structured as follows. The considered model selection criteria are described in Section~\ref{sec:eval_metrics}. Section~\ref{sec:monte_carlo} presents the design and the implementation details of the MC simulations. The results of the MC simulations for three different DGPs are presented in Section~\ref{sec:results}, which is divided in two subsections to address the main trade-off between \textit{number} of topics and \textit{coherence/structure} of the uncovered topics. The final section summarises the findings and provides recommendations for applications.

\section{Model Selection Criteria for LDA}\label{sec:eval_metrics}

The selection of the optimal  number of topics for LDA models can be based either on  measures of topic quality (similarity or coherence) or on measures of goodness-of-fit and model complexity.

Let us consider an LDA model under a standard ``bag-of-words'' assumption. For a document corpus $\mathcal{D}$ that consists of $J$ documents, each document $j$ ($j=1,2,\ldots,J$) is a set of $N_j$ words, where the ordering of words is ignored. The total number of words in the corpus is equal to $N=\sum_{j=1}^{J} N_j$. The document corpus $\mathcal{D}$ can be characterized by a $J\times I$ document-word frequency matrix  $X=\{x_{ji}\}_{j,i=1}^{J,I}$, where $x_{ji}$ is the frequency of word $i$ encountered in document $j$ and $I$ is the number of different words in the vocabulary.

Under the ``bag-of-words'' assumption, an LDA model can be summarized by a $J\times K$ matrix $\theta$ of document-topic frequencies and a $K\times I$ matrix $\beta$ of topic-word frequencies with the dimensions of these matrices depending on the number of topics $K$. The estimated document-term matrix is a product of estimates $\hat{\theta}$ and $\hat{\beta}$: $\hat{X}= \hat{\theta} \times \hat{\beta}$. A set of candidate LDA models is determined by the numbers of topics in candidate models: $K\in\{K_{min},\ldots,K_{max}\}$.

In the following, we describe two popular semantic measures of topic quality, which are often used in applications, and two recently developed goodness-of-fit measures.

\subsection{Topic Similarity}

Following~\citeasnoun{Cao2009}, the optimal number of topics is often selected by minimizing the average cosine similarity across topics:

\begin{equation*}
Cao\_Juan(K)=\frac{\sum_{k=1}^{K}\sum_{l=k+1}^{K}corr(k,l)}{K\times(K-1)/2},
\end{equation*}

\noindent where

\begin{equation*}
corr(k,l)=\frac{\sum_{i=1}^{I}\beta_{ki}\beta_{li}}{\sqrt{\sum_{i=1}^{I}\beta_{ki}^2}\sqrt{\sum_{i=1}^{I}\beta_{li}^2}},
\end{equation*}

\noindent and $\beta_{ki}$ is the frequency of word type $i$ in topic $k$.

The average cosine similarity is extensively used for selecting the number of topics in different text-as-data applications, e.g. analyzing scientific articles to examine the evolution of research over time and identify future fields of research~\cite{Loureiro2021,Tiba2018}, analyzing the speeches by Executive Board members of the European Central Bank~\cite{Hartmann2018}, investigating news data in the context of economic reforms~\cite{Lin2022}, analyzing and categorizing innovation projects~\cite{Dahlke2021}.

\subsection{Topic Coherence}

\noindent \citeasnoun{Mimno2011} proposed a model selection procedure that maximizes the average semantic coherence of topics:

\begin{equation*}
Mimno(K)= \frac{1}{K}\sum_{k=1}^{K} coh(k,\mathbf{i}^{(k)}),
\end{equation*}

\noindent where $coh(k,\mathbf{i}^{(k)})$ is the coherence metric for topic $k$,

\begin{equation*}
coh(k,\mathbf{i}^{(k)})=\frac{2}{v\times(v-1)}\sum_{m=2}^{v}\sum_{n=1}^{m-1}\log\frac{f(i_m^{(k)},i_n^{(k)})+\epsilon}{f(i_n^{(k)})},
\end{equation*}

\noindent $\mathbf{i}^{(k)}=(i_1^{(k)},\ldots,i_v^{(k)})$ is the list of the $v$ most frequent word types in topic $k$, $f(i)$ is the document frequency of word $i$ (i.e., the number of documents with at least one token of type $i$), and $f(i,i')$ is the co-document frequency of words $i$ and $i'$ (i.e., the number of documents containing one or more tokens of type $i$ and at least one token of type $i'$). The default number of the most probable words used is equal to 20. The smoothing parameter $\epsilon$ is included to avoid taking the logarithm of zero and its default value is equal to $e^{-12}$.

The average  semantic coherence is also often used for selecting the number of topics in applied topic mining analyzing, e.g., monetary policy speeches~\cite{Ferrara2022}, news data to forecast aggregated stock returns~\cite{Adammer2020}, energy market tweets with regard to their impact on market movements~\cite{Polizos2022}, or survey responses on the consequences of Covid-19 pandemic~\cite{Kleinberg2020}.

\subsection{OpTop Criterion}

\noindent~\citeasnoun{Grossetti2022} proposed to use a goodness-of-fit statistic based on the comparison of actual and estimated document-word frequencies.
The frequency of word type $i$ in document $j$ estimated in an LDA model with $K$ topics is

\begin{equation*}
\hat{x}^{(K)}_{ji}=\sum_{k=1}^K \hat{\theta}^{(K)}_{jk}\hat{\beta}^{(K)}_{ki}.
\end{equation*}

Because the matrix of document-word frequencies is usually sparse,~\citeasnoun{Grossetti2022} suggest collapsing relatively unimportant words in a single frequency bin. For document $j$, they order word types from the smallest to the largest estimated frequency, $(i^{(j)}_{1},i^{(j)}_{2},\ldots, i^{(j)}_{I})$ such that $\hat{x}^{(K)}_{ji_1}$ $\leq$ $\hat{x}^{(K)}_{ji_2}$ $\leq$ $\ldots$ $\leq$ $\hat{x}^{(K)}_{ji_I}$, and select a sub-vector of relatively unimportant word types $(i^{(j)}_{1},i^{(j)}_{2},\ldots, i^{(j)}_{p})$. The cumulative frequency of relatively unimportant word types in document $j$ estimated in an LDA model with $K$ topics is

\begin{equation*}
\hat{x}^{(K)}_{j,min}=\sum_{i\in(i^{(j)}_{1},\ldots,i^{(j)}_{p})}\hat{x}^{(K)}_{ji},
\end{equation*}

\noindent where  $\hat{x}^{(K)}_{ji_p}$ is the largest frequency such that $\sum_{i=i^{(j)}_1}^{i^{(j)}_p}\hat{x}^{(K)}_{ji}<x_{cutoff}$, and $x_{cutoff}$ is a cumulative frequency cut-off value. Following Lewis and Grossetti (2022), we use $x_{cutoff}=0.05$ as a baseline cut-off value. (For a robustness check, we also consider a cut-off value of $0.20$)  The resulting goodness-of-fit statistic is

\begin{equation}
OpTop(K)=\sum_{j=1}^{J}\left[(P_j+1)\left(\sum_{i\in(i^{(j)}_{p+1},\ldots,i^{(j)}_{I})}\frac{(\hat{x}^{(K)}_{ji}-x_{ji})^2}{\hat{x}^{(K)}_{ji}}+\frac{(\hat{x}^{(K)}_{j,min}-x_{j,min})^2}{\hat{x}^{(K)}_{j,min}}\right)\right], \label{OpTop}
\end{equation}

\noindent where $(i^{(j)}_{p+1},\ldots,i^{(j)}_{I})$ is a sub-vector of relatively important word types  in the $j$th document and $P_j$ is the length of this sub-vector.~\citeasnoun{Grossetti2022} propose to select an optimal number of topics by minimizing the OpTop statistic (\ref{OpTop}) over a range of numbers of topics. Unlike criteria
proposed by ~\citeasnoun{Cao2009} and \citeasnoun{Mimno2011},  the OpTop statistic is not a semantic measure of topic quality, but a goodness-of-fit measure that can be easily computed.

\bigskip

\subsection{Singular Bayesian Information Criterion}

The last model selection criterion -- singular Bayesian information criterion is a version of the Bayesian information criterion (BIC) which can be applied to singular statistical models \cite{Hayashi2020}, for which it is more suitable for model selection than BIC. The method was successfully applied by ~\citeasnoun{Bystrov2022} for selecting parsimonous LDA models with coherent topics, however the properties of the criterion as applied to LDA modelling have not been studied in a simulation setup.

Computation of sBIC is based on several results from the literature. The first one, is the decomposition of log-marginal likelihood of a text corpus $\mathcal{D}$ with $K$ topics, described by \citeasnoun{Watanabe2009}. In the context of an LDA model, this representation can be written as

\begin{center}\begin{math}
\log L(\mathcal{D}|K) = \log P(\mathcal{D}|\hat{\theta},\hat{\beta},K)- \lambda(K)\log(N)+(m(K)-1)\log\log(N)+O_p(1),
\end{math}\end{center}

\noindent where $\hat{\theta}$ and $\hat{\beta}$ are consistent estimators of the document-topic and topic-word probability matrices, respectively, $\lambda(K)$ is a learning coefficient, measuring stochastic complexity of a model with $K$ topics, $m(K)$ is its multiplicity and $N$ stands for the total number of words in the corpus.

In the next step, to deal with the problem of unknown values of $\lambda(K)$ and $m(K)$ which depend on the true value of $K$, model averaging described in \citeasnoun{Drton2017} is applied. In this approach, the singular Bayesian information criterion for an LDA model with $K$ topics can be defined as an approximation of the log-marginal likelihood obtained by averaging of models with smaller number of topics (see \citeasnoun{Drton2017}):

\begin{equation}\label{sbic}
sBIC(K)=\log L'(\mathcal{D}|K),
\end{equation}

\noindent where $L'(\mathcal{D}|K)$ denotes the approximation following from the \citeasnoun{Drton2017} procedure. To compute the marginal likelihood for every sub-model with the number of topics $k$ where $k\le K$, the formulas for the learning coefficient $\lambda(k)$ and its multiplicity $m(k)$ derived for LDA by \citeasnoun{Hayashi2021} are used.
Model selection is then based on maximizing the sBIC value.

As described by ~\citeasnoun{Bystrov2022} evaluation of~(\ref{sbic}) for some datasets may be associated with numerical problems resulting from very small values of the likelihood function. In order to avoid these problems, in the experiments we use high-precision computations.

\section{Monte Carlo Simulation}\label{sec:monte_carlo}

Despite the broad usage of the evaluation metrics described in the previous section, there is no consensus yet on which metric performs best, when it comes to choosing the number of topics. Given that the ground truth, i.e., the actual data generating process is unknown in real applications, the performance of the metrics can only be assessed based on a subjective analysis of the uncovered themes. To account for this problem, a Monte Carlo simulation study is required, for which the data are generated by a well defined DGP with a known number of different topics.\footnote{The idea of using Monte Carlo simulations for obtaining well-defined text corpora has been applied recently by \citeasnoun{Wang2021} in the context of model selection for text classification tasks. The authors use the generated data to evaluate the classification performance of different topic models.} This allows not only to compare the performance of alternative metrics with regard to the number of topics identified, but also to evaluate whether certain characteristics of the corpora such as number or length of documents might affect the relative performance. Furthermore, it enables to evaluate not only the number of topics identified, but also whether these topics correspond closely to the topics underlying the DGP, i.e., focusing also on the content in an objective approach.

This section provides the details of the Monte Carlo simulation setup used for the comparison of the methods described in Section~\ref{sec:eval_metrics}. First, in Subsection~\ref{section:procedure} we present the general framework that is applied for each of three different DGPs. Second, in Subsection~\ref{section:dgp} we describe the DGPs, which are derived from actual corpora with typical characteristics of textual data used in applications. Finally, Subsection~\ref{section:details} provides some technical implementation details.

\subsection{Procedure}
\label{section:procedure}

\begin{figure}[H]
    \centering
    \includegraphics[width=\textwidth]{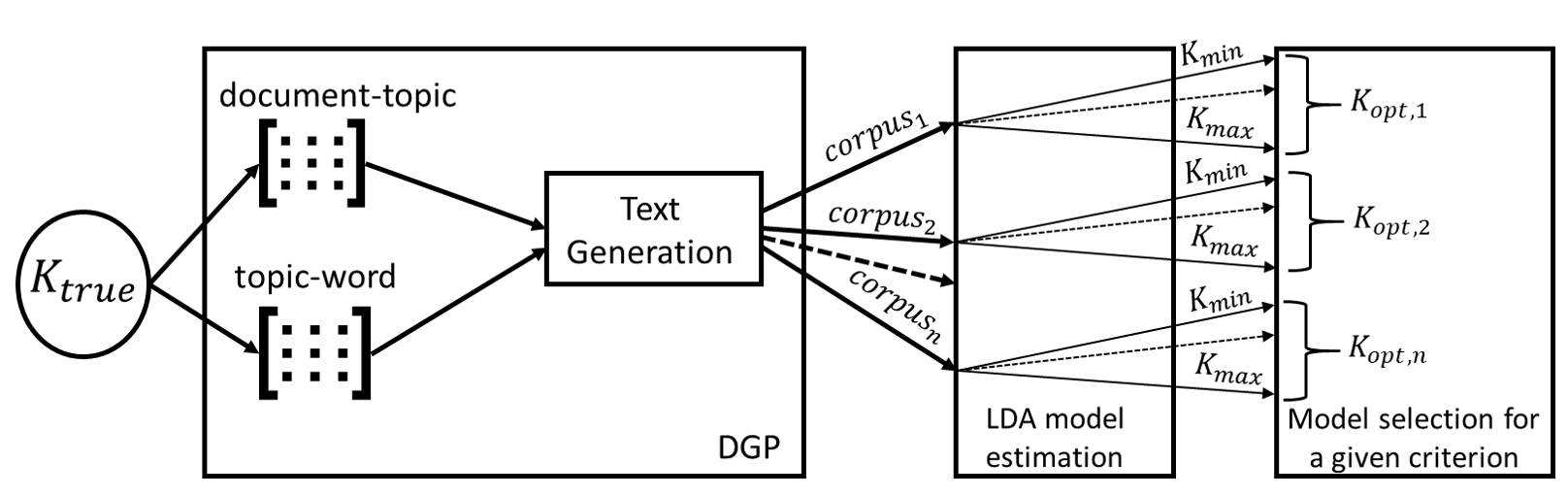}
    \caption{Generic procedure for Monte Carlo simulations with a given selection criterion}
    \label{fig:procedure}
\end{figure}

The three DGPs used in the Monte Carlo simulations are designed to replicate the characteristics of a given real document corpus. Figure~\ref{fig:procedure} presents the generic procedure which is applied to each of these DGPs.

As described in Section~\ref{sec:eval_metrics}, LDA is based on the assumption that each document in a corpus is a distribution over a given number $K_{true}$ of latent topics and each topic is a distribution over a fixed corpus vocabulary~\cite{Blei2003}. Thus, an LDA model is described by two matrices, the first containing the probabilities of occurrence of each word in each topic (topic-word distribution), and the second providing the probabilities of each topic occurring in a single document (document-topic distribution). The approach for generating Monte Carlo text corpora is based on these two matrices.

Therefore, in the first step of the procedure, an LDA model is estimated using a given real document corpus with a number of topics that was used in previous analysis of the selected corpus. In order to make sure that only distinct topics will be used for the generation of Monte Carlo text corpora in the following step, topics exhibiting a cosine similarity measure with other topics larger than a selected cut-off value (95\% or 99\% percentile) are dropped (see Appendix~\ref{Appendix_Topics} for more details). This correction allows to reduce a potential bias that can be induced by high topic similarity observed in the generated data and ensures a more stable performance of all model selection criteria.

For the remaining $K_{true}$ topics, the document-topic matrix is re-scaled to ensure that topic weights add up to one and passed on to the second step of text generation. The text generation process based on LDA is presented in Algorithm~\ref{alg1}. For each document in the original corpus, a new Monte Carlo document is created with the same number of words and document-topic distribution. For each word in this document, first, a topic is randomly selected based on the known document-topic distribution. Then, the word is randomly selected from the known vocabulary using the known topic-word distribution. It should be noted that the algorithm as presented does not exactly reproduce the generative procedure described in~\citeasnoun{Blei2003},  where document-topic and topic-word frequency matrices are obtained using hyper parameters. In applications, these hyper parameters are not often estimated, and using flat priors would result in distributions of document-topic and topic-word frequencies different from the ones actually observed. Therefore, we follow a data-driven approach and try to replicate closely the properties of the observed datasets serving as the benchmarks for our Monte Carlo corpora and, consequently,  use the estimated document-topic and topic-word frequency matrices.

\begin{algorithm}[H]
	\caption{Text generation\label{alg1}}
	\begin{algorithmic}[1]
		\For {$document=1,2,\ldots,J$}
			\State $document\_length = original\_document\_length$
			\For {$word=1,2,\ldots,document\_length$}
				\State \parbox[t]{11.5cm}{Randomly select a topic from the document-topic distribution \linebreak
        \hspace*{0.2cm} of the current document}\vspace*{0.2cm}
				\State Randomly select a word from the topic-word distribution
                \State Append the selected word to the current document
			\EndFor
			\State Append the generated document to the corpus.
		\EndFor
	\end{algorithmic}
\end{algorithm}

Algorithm~\ref{alg1} is applied to each DGP with 500 
Monte Carlo replications, i.e., 500 
corpora containing the same number of documents of same length as the original corpus.

In the third step, for each criterion that we use, we estimate the LDA model with number of topics within the interval $[ \max\{2; K_{true} - 20\}, K_{true} + 20]$, where $K_{true}$ is the number of topics used when generating text corpora from the DGPs. The maximum length of the range of admitted values for the number of topics is equal to 40 with the true number of topics $K_{true}$ of the DGP in the center of the interval if larger than 20. Otherwise, the lower bound is set to 2, the lowest sensible number of topics. This limited range of admitted values for the number of topics is due to the high computational costs of model estimation. The optimal number of topics is determined for each of the selected criteria based on these estimated models.

For the final step, the comparison of the outcomes of different model selection criteria in Subsection~\ref{sec:topic_number}, we use descriptive statistics such as standard deviation, mean, median, and skewness. For the visualisation of the distributions over the number of topics determined according to the considered criteria, we use histograms. Furthermore, in Subsection~\ref{sec:topic_content}, we will also provide information about the extent to which the topics used for generating the texts are found when applying LDA with the number of topics selected by the different criteria.

\subsection{Data Generation Processes\label{dgp}}
\label{section:dgp}

The three DPGs used for the Monte Carlo simulations are related to three real world corpora:

\begin{itemize}
    \item DGP 1 replicates the characteristics of a corpus consisting of scientific papers published in the \href{https://www.degruyter.com/view/journals/jbnst/jbnst-overview.xml}{Journal of Economics and Statistics} (JES).
    \item DGP 2 reproduces features of the corpus consisting of abstracts submitted to \href{https://www.ercim.eu/}{European Research Consortium for Informatics and Mathematics (ERCIM)} and \href{http://www.cfenetwork.org/}{Computational and Financial Econometrics (CFE)} conferences.
    \item DGP 3 reproduces the properties of a corpus containing Newsticker items from \href{https://www.heise.de/}{heise online}.
\end{itemize}

The data from JES used for DGP 1 cover the period from 1984 to 2020 and consists of 704 documents with an average text length of about 3,000 words. The size of the vocabulary for this corpus is equal to 3,911 words. The collection focuses on scientific publications in empirical economics and applied statistics. The initial number of topics selected was equal to 60 as in \citeasnoun{Bystrov2022}. After removing topics which were too similar, the final number of topics used in DGP 1 is equal to 38 ($K_{true} = 38$).

The conference abstract data used for DGP 2 cover the period from 2007 to 2019 and consists of 11,387 documents with an average text length of about 80 words. For this corpus the dictionary is composed of 1,796 words. The focused nature of conference abstracts allows to expect a limited number of topics. The initial number of topics selected for these data was equal to 20. This number is reduced to 12 ($K_{true} = 12$) after removing the topics that were too close to each other.

The heise data used for DGP 3 cover the period from 1996 to 2021 and include 181,402 documents with an average length of about 120 words. The number of words in the vocabulary for this corpus is equal to 4,675. The news platform discusses a significant number of topics concerning technological advances. The initial number of topics selected was equal to 120. Keeping the most distinct topics, the final number of topics used in DGP 3 was equal to 70 ($K_{true} = 70$). In the analysis we use only the most recent 50,000 documents from this corpus because using the whole dataset would increase the computational costs for the Monte Carlo simulation beyond the available capacities.

\subsection{Details of Implementation\label{implementation}}
\label{section:details}

All Monte Carlo simulations were implemented using Python. To generate random sequences used in the text generation stage (Algorithm~\ref{alg1}), the random number generator from Pythons' numpy package was used (\url{https://numpy.org/doc/stable/reference/random/generator.html}).
LDA models were estimated using the Gibbs sampler as implemented in the Python package "lda" (\url{https://pypi.org/project/lda/}). For each corpus generated based on DGP 1, models with topic numbers in the interval $[18;58]$ were estimated, for corpora generated based on DGP 2 - in the interval $[2;32]$, and for corpora generated based on DGP 3 - in the interval $[50;90]$. Most other parameters of the package were used at the default values. The number of iterations was set to a relatively small value of 1\,000 due to computational constraints.

The Cao\_Juan and Mimno criteria were computed using the Python package "tmtoolkit" (\url{https://pypi.org/project/tmtoolkit/}).
The Python implementations of the sBIC and OpTop model selection criteria were written by the authors. Computations were performed using the high-performance-computing-cluster at Justus Liebig University Giessen (justHPC) (\url{https://www.hkhlr.de/de/cluster/justhpc-giessen}).\footnote{Code details can be found in the Github repository for this paper at \url{https://github.com/VikaNa/sBIC}
.}

\section{Results}\label{sec:results}

This section summarizes the results of the Monte Carlo simulations. It is divided into two subsections. The first (\ref{sec:topic_number})
presents and discusses the results on finding the optimal (true) number of topics. The second one (\ref{sec:topic_content}) focuses on topics' contents and structure, and proposes a procedure to analyse/evaluate both.

\subsection{Number of Topics}\label{sec:topic_number}

\begin{figure}[H]
\centering
\includegraphics[width=\textwidth]{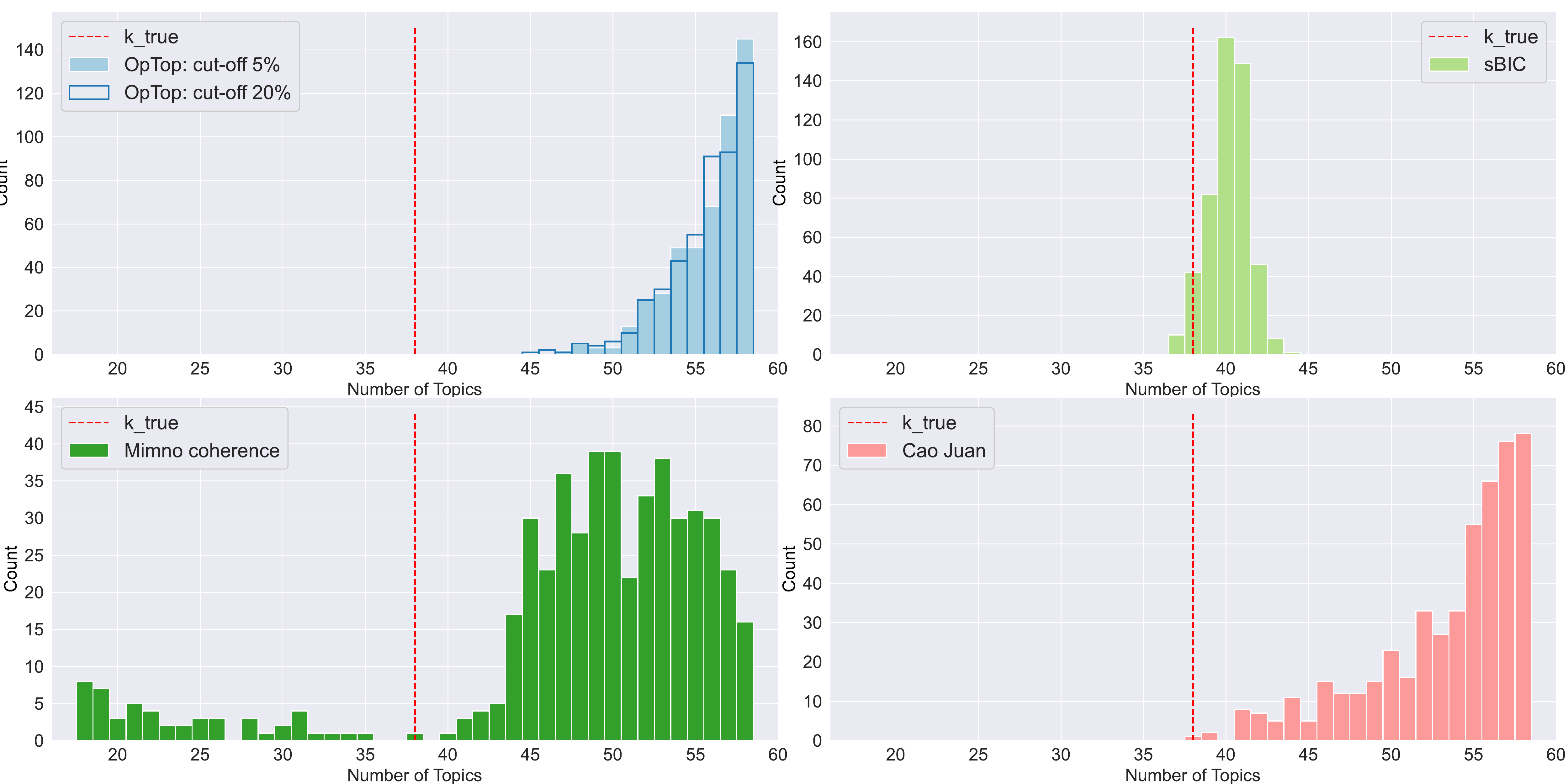}
\caption{Comparison of evaluation metrics for DGP1}
\label{fig: dgp1_comparison}
\end{figure}

\begin{figure}[H]
\centering
\includegraphics[width=\textwidth]{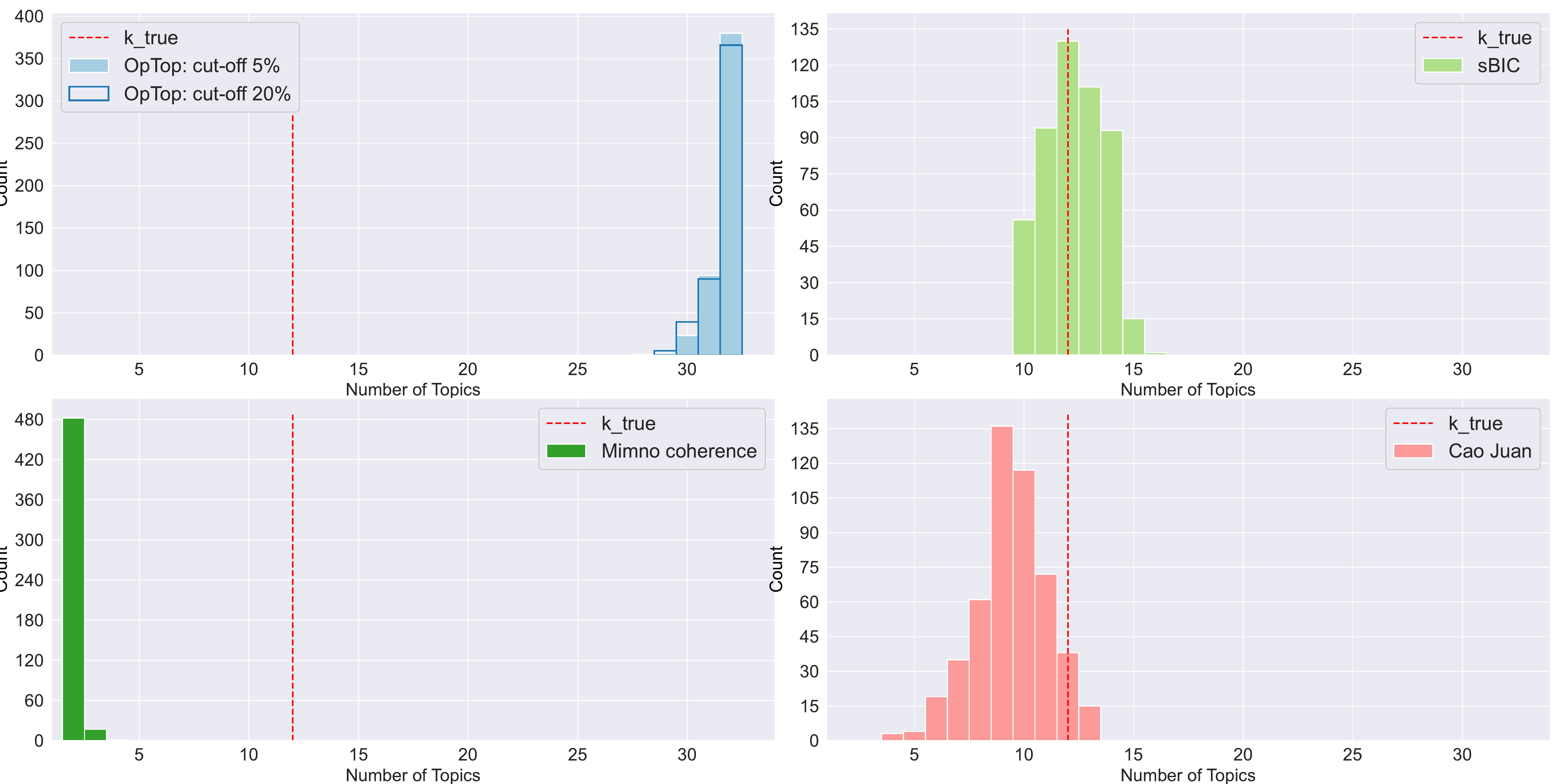}
\caption{Comparison of evaluation metrics for DGP2}
\label{fig: dgp2_comparison}
\end{figure}

\begin{figure}[H]
\centering
\includegraphics[width=\textwidth]{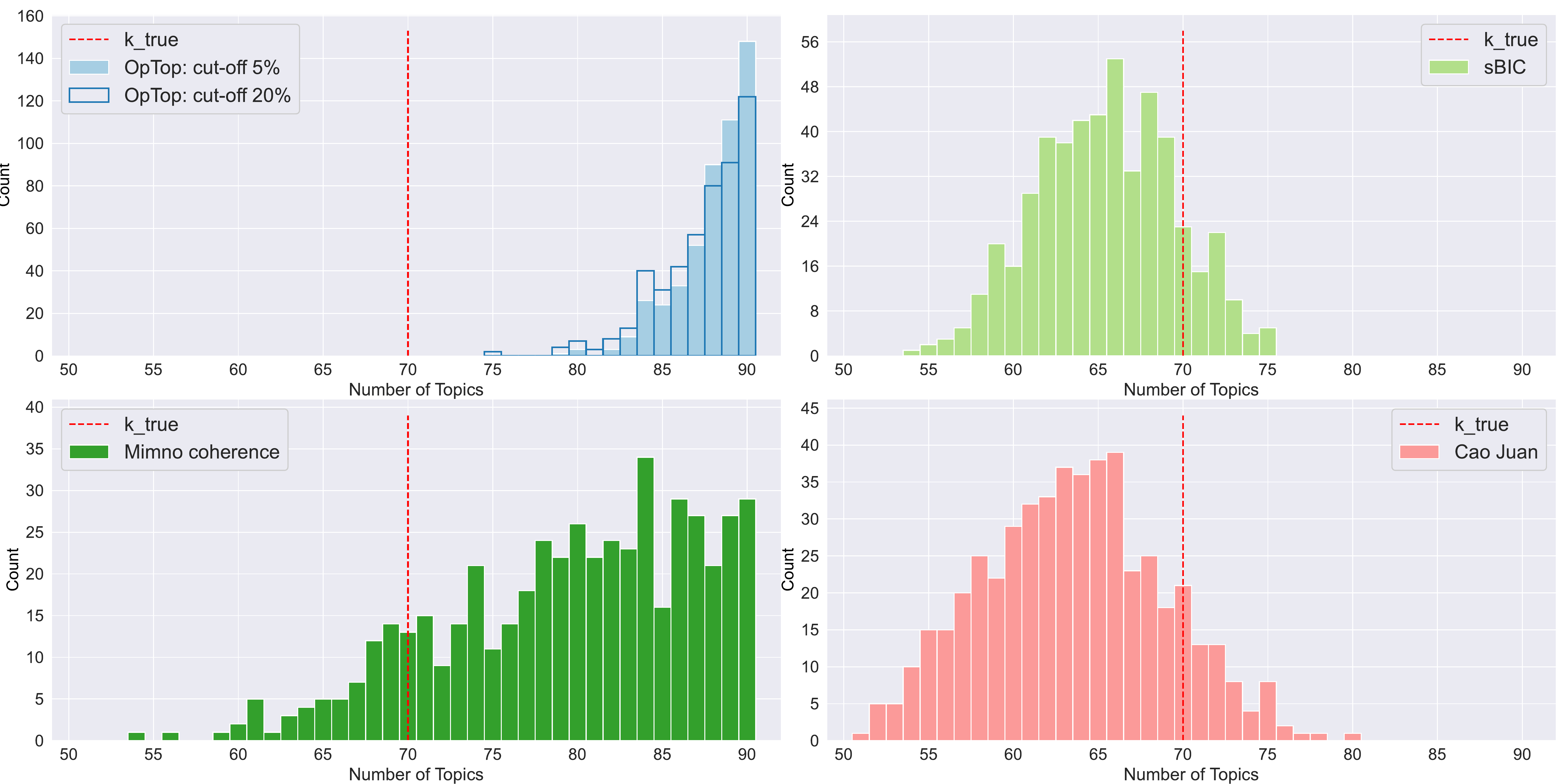}
\caption{Comparison of evaluation metrics for DGP3}
\label{fig: dgp3_comparison}
\end{figure}

The first set of results concerns the estimation of the number of topics $K$.
Figures~\ref{fig: dgp1_comparison},~\ref{fig: dgp2_comparison} and \ref{fig: dgp3_comparison} present histograms for the numbers of themes selected by the evaluation metrics described in Section~\ref{sec:eval_metrics} for all three considered DGPs. In each of the histograms, the red vertical line depicts the true number of topics ($K_{true}$) used for generating the corpora. Table~\ref{tab: statistics} provides selected descriptive statistics computed for these estimates. The shape and location of histograms shown in Figures 2-4 suggest that sBIC is clearly the best method for selecting the number of topics for DGP 1 and DGP 2, while it performs similarly to the method of~\citeasnoun{Cao2009} for DGP 3.

\captionsetup[figure]{font=small,skip=0pt}

Statistics from Table 1 further indicate that the mean and the median of the estimated number of topics for sBIC is in fact the closest to the true value for all DGPs. For DGP 2 the median of the estimates provided by sBIC is the actual number of themes. The performance of the metric differs for DGP 1 and DGP 3. In the first case, sBIC tends to select too many topics and in the second it chooses too few topics on average. The differences between the true and the estimated values are relatively small, however both types of estimation errors have their consequences. Overestimation means that some artificial topics will be generated, while underestimation implies that a number of relevant themes will be omitted. These issues are further discussed in Section~\ref{sec:topic_content} where the structure and content of the estimated topics is evaluated.

The performance of the OpTop statistic is rather poor as the procedure has a strong tendency to select too many topics for each DGP for both cut-off values of low-frequency words (5\% and 20\%). In each case, the mean and median values of the estimates are very close to the maximum of the range of candidates for the optimal number of topics. Such large overestimation errors mean that a substantial number of topics that do not belong would be estimated. Since, as noted by \citeasnoun{Mimno2011}, there is a trade-off between obtaining many refined topics and meaningful themes, the quality of these additional topics found by the OpTop method might be expected to be rather low.

The working of the average cosine similarity (Cao\_Juan) depends on the DGP. The mean/median number of topics selected for DGP 1 is too large as compared with the true number of topics, while the mean/median number of topics selected for DGPs 2 and 3 is too low as compared to the true number of topics. This outcome might depend on particular features of the DGPs (e.g. DGP 1 including a relatively small number of longer documents) which could be subject to further analyses. On the whole, the estimation errors are larger than for sBIC and smaller than in case of the OpTop procedure.

The unsystematic behaviour in terms of the tendency to over- or underestimate can be also seen for the average semantic coherence (Mimno). The mean/median number of topics selected for DGP 1 and DGP 3 are too large as compared with the true number of topics, while there is severe underestimation problem for DGP 2. The performance of this procedure seems to be quite unstable as in the case of DGPs 1 and 3 the estimates have the largest variance as compared to the remaining methods.
\begin{table}[H]
\begin{center}
  \begin{tabular}{|l|l|r|r|r|}
\toprule \hline
          &  &   \makecell{DGP1 \\($K_{true}=38$)}  &  \makecell{DGP2 \\($K_{true}=12$)} &   \makecell{DGP3 \\($K_{true}=70$)}\\\cline{1-5}
\midrule
\multirow{4}{*}{sBIC} & std &   1.23 &   1.35 &   4.09 \\ \cline{2-5}
          & mean &  40.15 &  12.28 &  65.43 \\ \cline{2-5}
          & median &  40.00 &  12.00 &  66.00 \\ \cline{2-5}
          & skewness &  -0.24 &   0.00 &  -0.04 \\ \cline{2-5}
\cline{1-5}
\multirow{4}{*}{Cao\_Juan} & std &   4.54 &   1.68 &   5.36 \\ \cline{2-5}
          & mean &  53.40 &   9.43 &  63.53 \\ \cline{2-5}
          & median &  55.00 &   9.00 &  64.00 \\ \cline{2-5}
          & skewness &  -1.16 &  -0.28 &   0.12 \\ \cline{2-5}
\cline{1-5}
\multirow{4}{*}{Mimno} & std &   9.23 &   0.20 &   7.55 \\ \cline{2-5}
          & mean &  47.88 &   2.04 &  79.50 \\ \cline{2-5}
          & median &  50.00 &   2.00 &  81.00 \\ \cline{2-5}
          & skewness &  -1.89 &   5.55 &  -0.61 \\ \cline{2-5}
\cline{1-5}
\multirow{4}{*}{OpTop 5\%} & std &   2.30 &   0.59 &   2.06 \\ \cline{2-5}
          & mean &  55.81 &  31.70 &  88.03 \\ \cline{2-5}
          & median &  57.00 &  32.00 &  89.00 \\ \cline{2-5}
          & skewness &  -1.22 &  -2.17 &  -1.28 \\ \cline{2-5}
\cline{1-5}
\multirow{4}{*}{OpTop 20\%} & std &   2.39 &   0.67 &   2.61 \\ \cline{2-5}
          & mean &  55.67 &  31.63 &  87.38 \\ \cline{2-5}
          & median &  56.00 &  32.00 &  88.00 \\ \cline{2-5}
          & skewness &  -1.37 &  -1.78 &  -1.30 \\ \cline{1-5}
\bottomrule
\end{tabular}
  
\end{center}
\caption{Evaluation of different criteria}
\label{tab: statistics}
\end{table}

\subsection{Content and Structure of Topics}\label{sec:topic_content}

While the selected \textit{number} of topics delivers first general insights on the performance of different criteria, this indicator does not contain information on the correspondence between the topics used to generate the text corpora and the topics obtained using the selected number of topics in the estimation procedure. Therefore, the structure and the \textit{content} of topics should be also considered.\footnote{In applications, sometimes the quality of topics is analyzed based on human judgment. For example, \citeasnoun{Morstatter2018} present an approach based on existing measures of topic coherence and extending them by a measure of topic consensus by humans. Although this approach delivers some measure of interpretability by humans, the authors point out the need for automated and reproducible measures of topic quality.} To this end, we propose to consider the problem as a classification task. This allows to compare the results obtained using all the different selection criteria quantitatively making use of well established performance metrics. We use precision and recall as commonly used performance metrics. In standard applications, these are defined as follows:
\begin{itemize}
    \item \textbf{Recall} describes how many relevant items are retrieved.
    \item \textbf{Precision} indicates how many retrieved items are relevant.
\end{itemize}

In standard classification tasks, the length of predicted and actual labels is the same. In our case it might be different, as the number of topics selected by each of the considered evaluation metrics can deviate from the true number of topics as described in the previous subsection. Thus, we define the True Positive (TP) class as those topics that were correctly identified, i.e., true topics which find their match in the set of estimated topics for the number of topics indicated by the given selection criterion. Using this definition, precision and recall can be defined and calculated as follows:
\begin{equation}
    \text{Recall}=\frac{|\text{TP}|}{K_{true}},
\end{equation}

\noindent where $|\text{TP}|$ denotes the cardinality of the set TP and
$K_{true}$ is the true number of topics in a particular DGP.

\begin{equation}
    \text{Precision}=\frac{|\text{TP}|}{K_{metric}},
\end{equation}

\noindent where $K_{metric}$ is the proposed number of topics according to the selection criterion considered.

As there might be a trade-off between recall and precision, the F1 measure is often used as a combined measure. F1 is calculated as follows:

\begin{equation}\label{eq_f1}
    \text{F1}=2*\frac{\text{Precision}*\text{Recall}}{\text{Precision}+\text{Recall}}.
\end{equation}

For computing these measures, estimated topics have to be matched with true topics from the DGP. This matching can be done using topic matching technique proposed by~\citeasnoun{Bystrov2022}, the so-called \textit{best matching}. For each ``true'' topic, a match in the set of estimated topics is identified using the cosine similarity measure. If one of the ``true'' topics finds several matches, we only consider the matches with the highest cosine similarities. Obviously, a ``best match'' does not have to be a sensible match, i.e., close to the true topic. Therefore, we apply a threshold for the cosine similarity which has to be surpassed in order to consider a match as being a sensible match. This threshold is the same as used for the topic number reduction step for each DGP described in Subsection~\ref{dgp} (see Appendix~\ref{Appendix_Topics} for further details).

Table~\ref{tab: recall_best_match} describes the distribution of precision and recall for each DGP and each evaluation metric.\footnote{As a robustness check we also calculate the described performance metrics using cosine similarities instead of the binary indicator match/no match. The procedure is described in Appendix~\ref{Appendix_WeightedRecall}. The results do not differ qualitatively.}
As mentioned before, our application differs from standard classification problems as the number of true and estimated topics might differ. Hence, the interpretation of the results is slightly different. Here, a precision value of 1 means that all of the estimated topics are sensible matches to some of the true topics. However, it does not imply that all of the true topics are uncovered. Consequently, this measure might overestimate the performance of a metric if it tends to underestimate the true number of topics. For DGP 2, for example, the Mimno metric is described by an average precision value of 1, while the average recall value is 0.17. In the previous subsection, it was shown that the Mimno metric tends to underestimate the true number of topics for DGP 2. Thus, the high precision value only indicates that these few estimated topics are related to the true topics. sBIC, on the other hand, shows relatively high values for both recall and precision, 0.93 and 0.92 respectively indicating that mostly true topics and most of the true topics are found.

As for recall, a value of 1 means that all of the true topics are uncovered by the estimated topics. However, it does not imply that $K_{metric}=K_{true}$. Consequently, this measure might lead to overestimation of the performance of a metric  if it tends to select too many topics. For DGP 1, for example, the Cao\_Juan metric reveals an average recall value of 1, while the average precision value of 0.72 is substantially lower. Also in this example, sBIC performs well with average recall and precision values of 0.99 and 0.94, respectively.

To take account of the trade-off described above, it seems appropriate to consider both evaluation metrics simultaneously. This is done making use of the F1 score defined in equation~\ref{eq_f1} a the harmonic mean of recall and precision. The interpretation of F1 is straightforward: the higher the values the better the joint score for both recall and precision. The results indicate that sBIC outperforms the other evaluation metrics for DGP 1 and DGP 2. For DGP 3, according to the F1 score sBIC is found to perform similarly to the Cao\_Juan criterion, while still exhibiting some advantages compared to the other criteria.

\begin{table}[]
    \centering
\begin{tabular}{|l|l|r|r|r|r|r|r|}
\toprule \hline
     &      & \multicolumn{2}{l|}{Recall} & \multicolumn{2}{l|}{Precision} & \multicolumn{2}{l|}{F1} \\ \cline{1-8}
  data   &   metric   &   mean &   std &      mean &   std &  mean &   std \\ \cline{1-8}
\midrule
\multirow{5}{*}{DGP1} & Cao\_Juan &   1.00 &  0.00 &      0.72 &  0.07 &  0.83 &  0.04 \\ \cline{2-8}
     & Mimno &   0.96 &  0.12 &      0.78 &  0.09 &  0.85 &  0.06 \\ \cline{2-8}
     & OpTop 20\% &   1.00 &  0.00 &      0.68 &  0.03 &  0.81 &  0.02 \\ \cline{2-8}
     & OpTop 5\% &   1.00 &  0.00 &      0.68 &  0.03 &  0.81 &  0.02 \\ \cline{2-8}
     & sBIC &   0.99 &  0.01 &      0.94 &  0.03 &  0.97 &  0.01 \\ \cline{2-8}
\cline{1-8}
\multirow{5}{*}{DGP2} & Cao\_Juan &   0.78 &  0.13 &      1.00 &  0.02 &  0.87 &  0.09 \\ \cline{2-8}
     & Mimno &   0.17 &  0.02 &      1.00 &  0.00 &  0.29 &  0.02 \\ \cline{2-8}
     & OpTop 20\% &   1.00 &  0.00 &      0.38 &  0.01 &  0.55 &  0.01 \\ \cline{2-8}
     & OpTop 5\% &   1.00 &  0.00 &      0.38 &  0.01 &  0.55 &  0.01 \\ \cline{2-8}
     & sBIC &   0.93 &  0.06 &      0.92 &  0.07 &  0.92 &  0.04 \\ \cline{2-8}
\cline{1-8}
\multirow{5}{*}{DGP3} & Cao\_Juan &   0.87 &  0.05 &      0.96 &  0.03 &  0.91 &  0.02 \\ \cline{2-8}
     & Mimno &   0.93 &  0.04 &      0.83 &  0.05 &  0.87 &  0.02 \\ \cline{2-8}
     & OpTop 20\% &   0.98 &  0.01 &      0.79 &  0.03 &  0.87 &  0.02 \\ \cline{2-8}
     & OpTop 5\% &   0.98 &  0.01 &      0.78 &  0.02 &  0.87 &  0.02 \\ \cline{2-8}
     & sBIC &   0.88 &  0.04 &      0.95 &  0.03 &  0.91 &  0.02 \\ \cline{1-8}
\bottomrule
\end{tabular}
    \caption{Descriptive statistics of recall, precision, and F1 scores}
    \label{tab: recall_best_match}
\end{table}

\section{Conclusions and Outlook}\label{conclusion}

Estimating LDA models requires making a number of decisions regarding parameter settings. This paper considered the problem of selecting the value of one of those essential parameters, viz.\ the number of topics discussed in the text corpus. The main aim was to analyze the properties of various model selection criteria with special focus on the recently proposed singular Bayesian information criterion. The performance of the methods was examined via Monte Carlo experiments using synthetic data generating processes based on empirical text corpora which differed with respect to the number and length of documents and the number of topics. The performance of different model selection procedures was evaluated by not only examining the accuracy of estimating the actual number of topics but also by analyzing the structure and contents of the estimated topics.

Simulation results showed that the singular Bayesian information criterion performed relatively well for all data generating processes considered in the experiments. It was the best method for estimating the number of topics as it was associated with the smallest estimation errors as compared to the competitors. In addition, it resulted in topics with good content and structure and performed in a relatively stable fashion for all data generating processes. Across the DGPs, the working of the method based on sBIC was worst for DGP 3 corresponding to a text corpus with a large number of short documents and a substantial number of topics. In this setting, sBIC exhibited a certain downward bias in the selected number of topics which might be taken into account in applied work. The reasons for this finding and possible adjustments to the method might be subject to further analyses.

The performance of the methods proposed by \citeasnoun{Cao2009} and \citeasnoun{Mimno2011} depended on the DGP. For each of these methods, the experiments revealed cases of systematic under- or overestimation of the true number of topics. The estimation errors were larger than those found for sBIC and had some negative consequences for the structure and content of the estimated topics. Dependence on the DGP implies that reliability and stability of these methods cannot be guaranteed in applied work unless further analyses will explain the relation between features of a DGP and the model selection results. Despite these drawbacks, the method of \citeasnoun{Cao2009} was still overall the second best approach to LDA model selection in the experiments reported in this paper. It was found that the method could be particularly useful for modelling collections of many short texts related to a large range of topics.

The final set of conclusions relates to the OpTop criterion. It was shown that the method tends to select models with an excessively large number of topics. The estimation errors were very substantial and led to small precision and F1 metric values used for examining the content and structure of estimated topics. These results imply that using this criterion in applied work can result in obtaining some spurious topics, which do not correspond to the data generating process. It seems that poor estimation properties of the OpTop procedure could be improved by the introduction of an appropriate penalty for model complexity (which increases with the number of topics) into the test statistic formula. This adjustment constitutes a further direction of future research.

\clearpage

{
    \bibliography{literature}}
\clearpage
\begin{appendices}
\section{Topic Number Reduction}\label{Appendix_Topics}

The goal of the topic number reduction step in preparing our DGPs for the Monte Carlo simulation was to use well separated topics allowing for a robust comparison of the topics estimated with the underlying DGPs. The process of topic number reduction comprises the following three steps:
\begin{enumerate}
    \item Starting with the estimated LDA for a given corpus, for each topic the most similar other topic is identified using the standard matching proposed by \citeasnoun{Bystrov2022}.
    \item For deciding whether a pair of topics is ``too similar'', i.e., will be excluded before generating synthetic data within the Monte Carlo simulation, a threshold value has to be defined. This value is also obtained by a data driven approach. We calculate all pairwise cosine similarity scores for each DGP providing $\frac{K^2-K}{2}$ typical values. Sorting them in increasing order provides the distributions shown in Figure~\ref{fig: topic_reduction}. Following the approach of the ``elbow'' criterion, we set percentile values defining the cut-off value for each DGP. These values are shown in the figure by the red horizontal line and correspond to the 95\% percentile for DGP2 and to the 99\% percentile for DGPs 1 and 3, respectively.
     \begin{figure}[ht]
    \centering
    \begin{subfigure}[t]{0.32\textwidth}
       \includegraphics[width=1\linewidth]{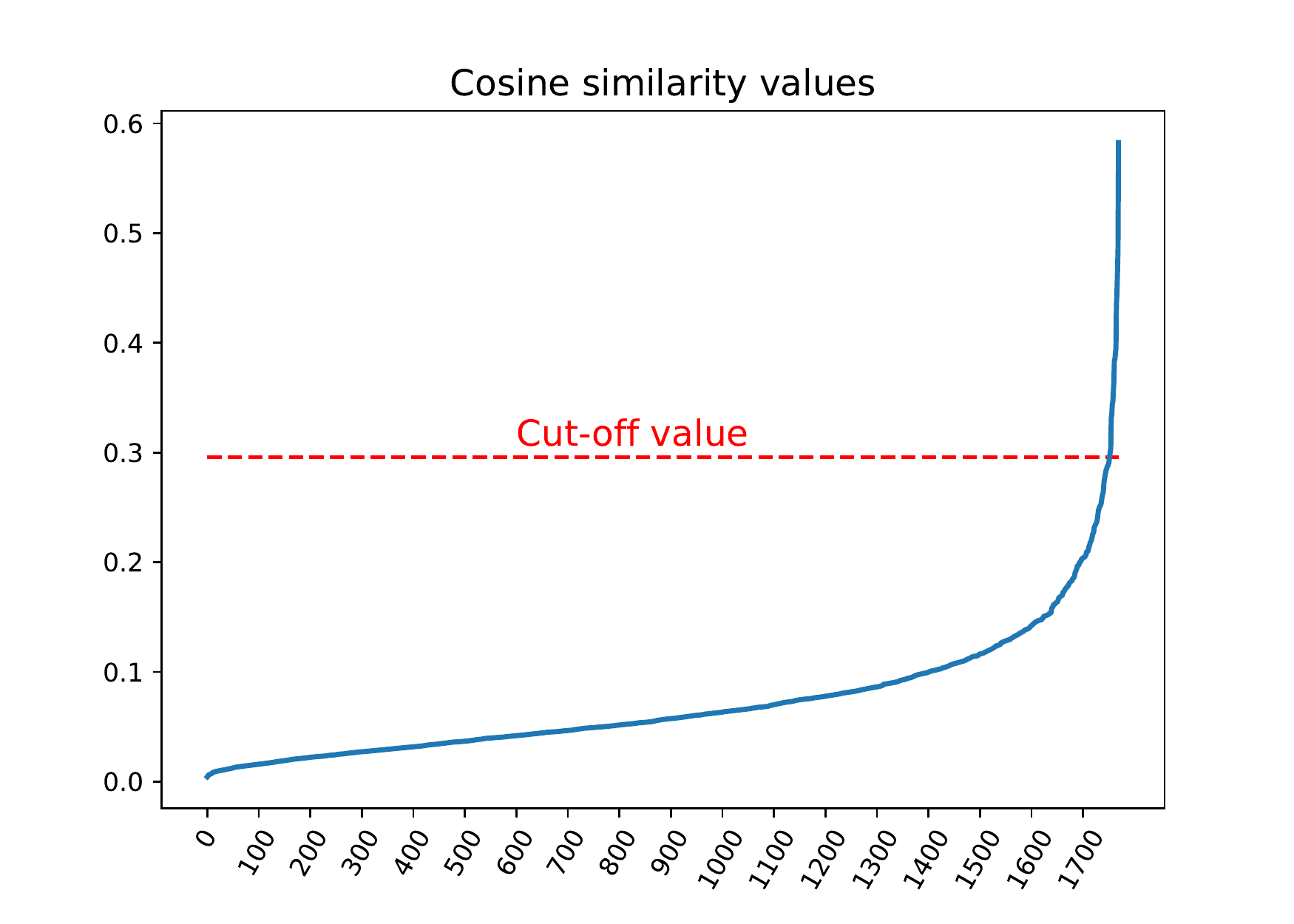}
       \caption{DGP1}
    \end{subfigure}
    \begin{subfigure}[t]{0.32\textwidth}
       \includegraphics[width=1\linewidth]{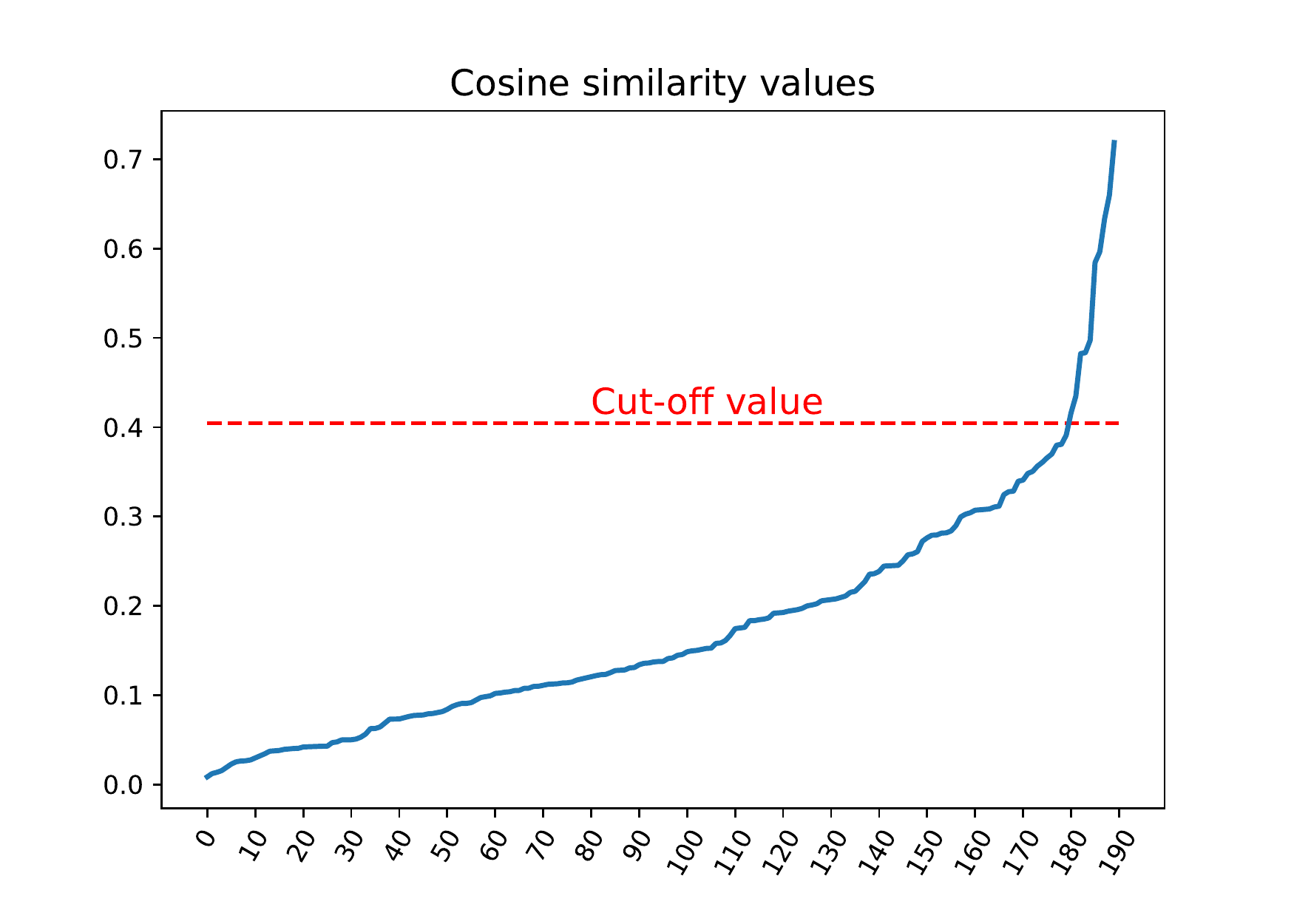}
       \caption{DGP2}
    \end{subfigure}
    \begin{subfigure}[t]{0.32\textwidth}
       \includegraphics[width=1\linewidth]{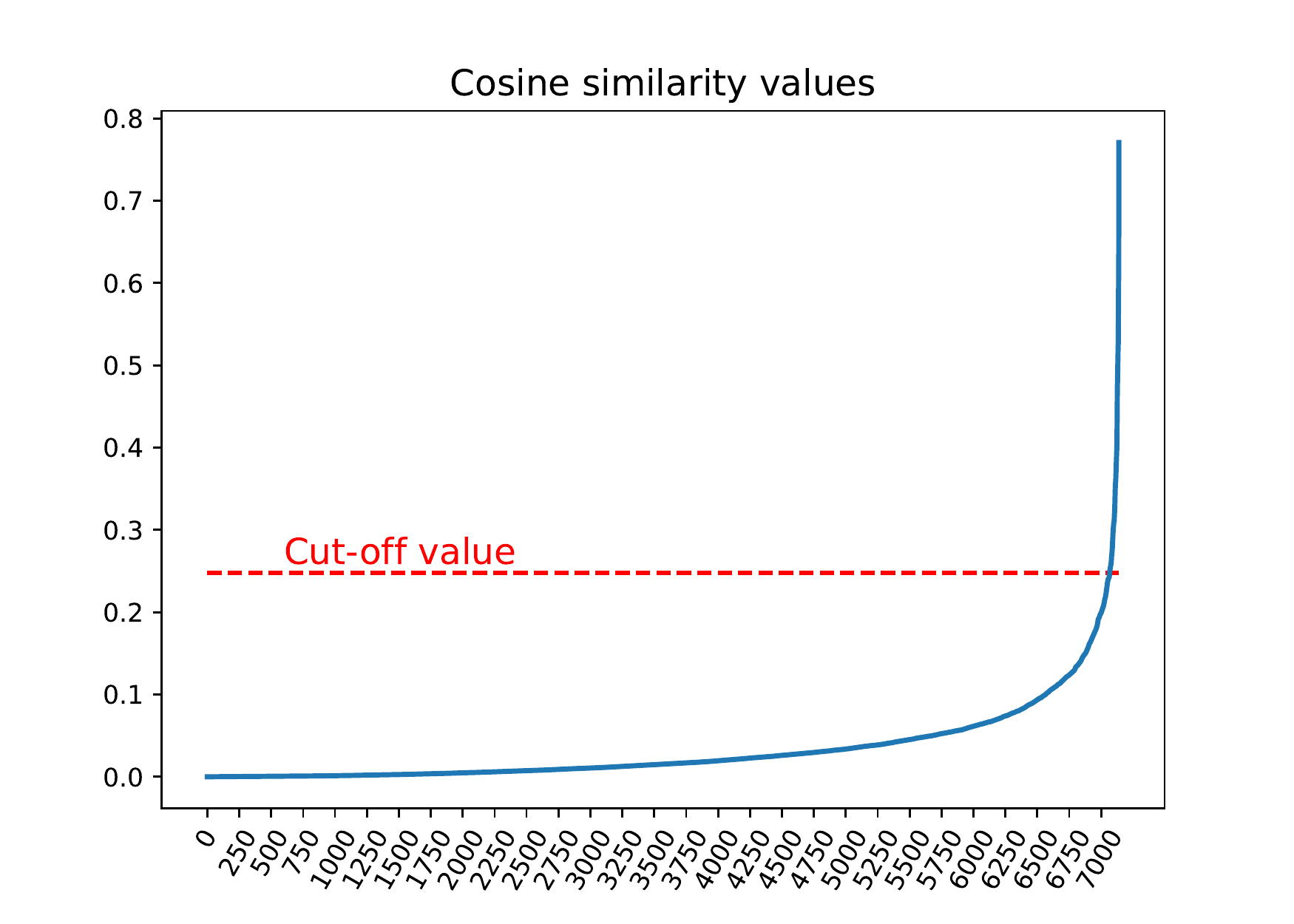}
       \caption{DGP3}
    \end{subfigure}
    \vspace*{0.3cm}

    \caption{Distribution of the pairwise cosine similarity values.}
    \label{fig: topic_reduction}
    \end{figure}
    \item All topics belonging to matched topic pairs above the cut-off value are considered as being too similar and, consequently, are removed from the model before starting the data generation within the Monte Carlo simulation. Figures~\ref{fig:similar_topics_dgp1},~\ref{fig:similar_topics_dgp2}, and~\ref{fig:similar_topics_dgp3} show examples of pairs including redundant topics in each DGP, which are eliminated by this method.
     \begin{figure}[H]
        \centering
        \includegraphics[width=0.9\linewidth]{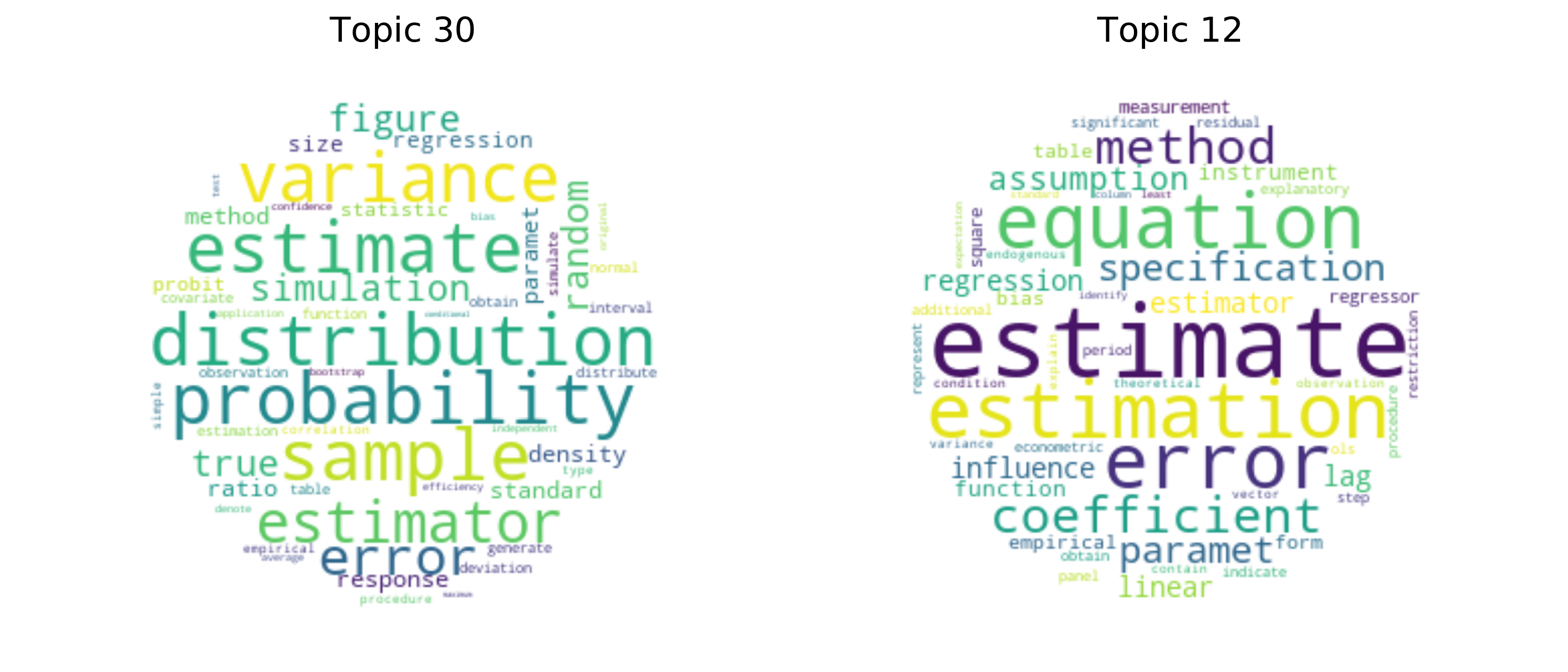}
        \caption{Similar topics in DGP 1}
        \label{fig:similar_topics_dgp1}
    \end{figure}
    \begin{figure}[H]
        \centering
        \includegraphics[width=0.9\linewidth]{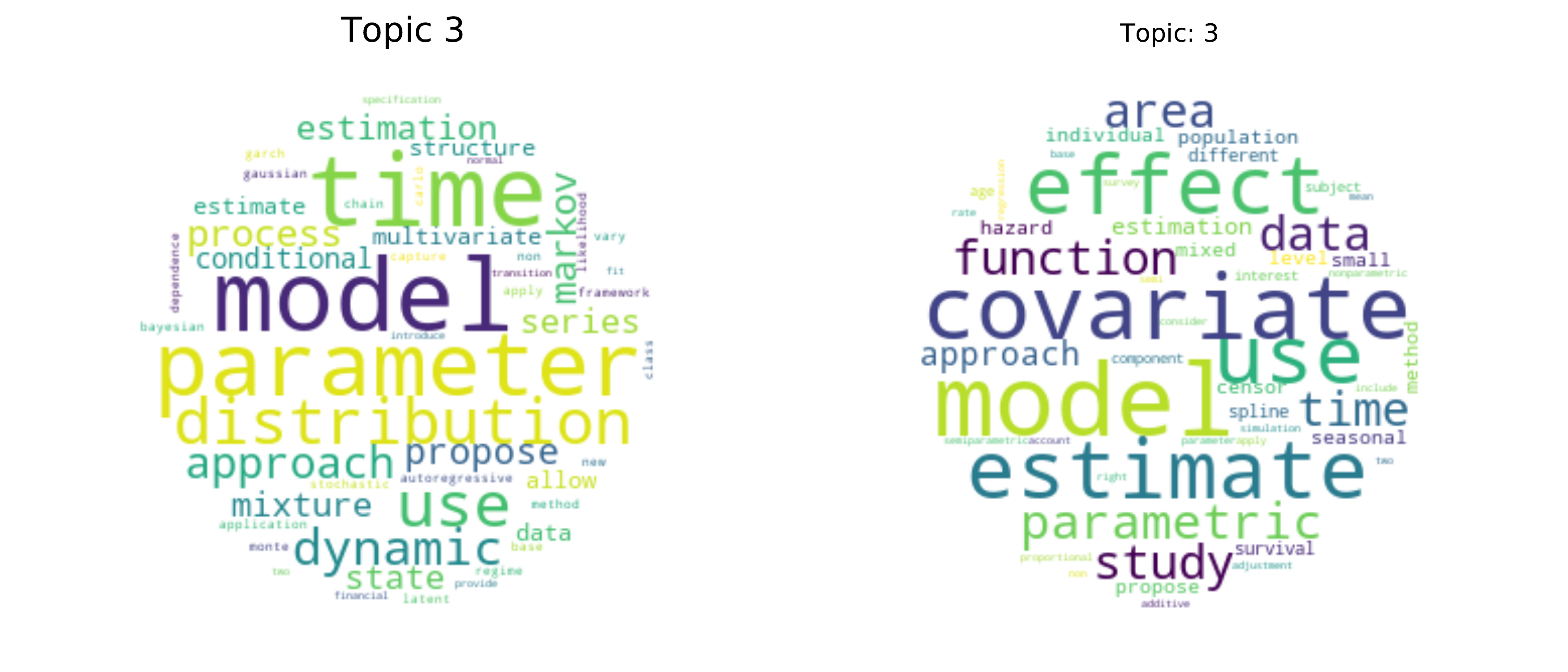}
        \caption{Similar topics in DGP 2}
        \label{fig:similar_topics_dgp2}
    \end{figure}
    \begin{figure}[H]
        \centering
        \includegraphics[width=0.9\linewidth]{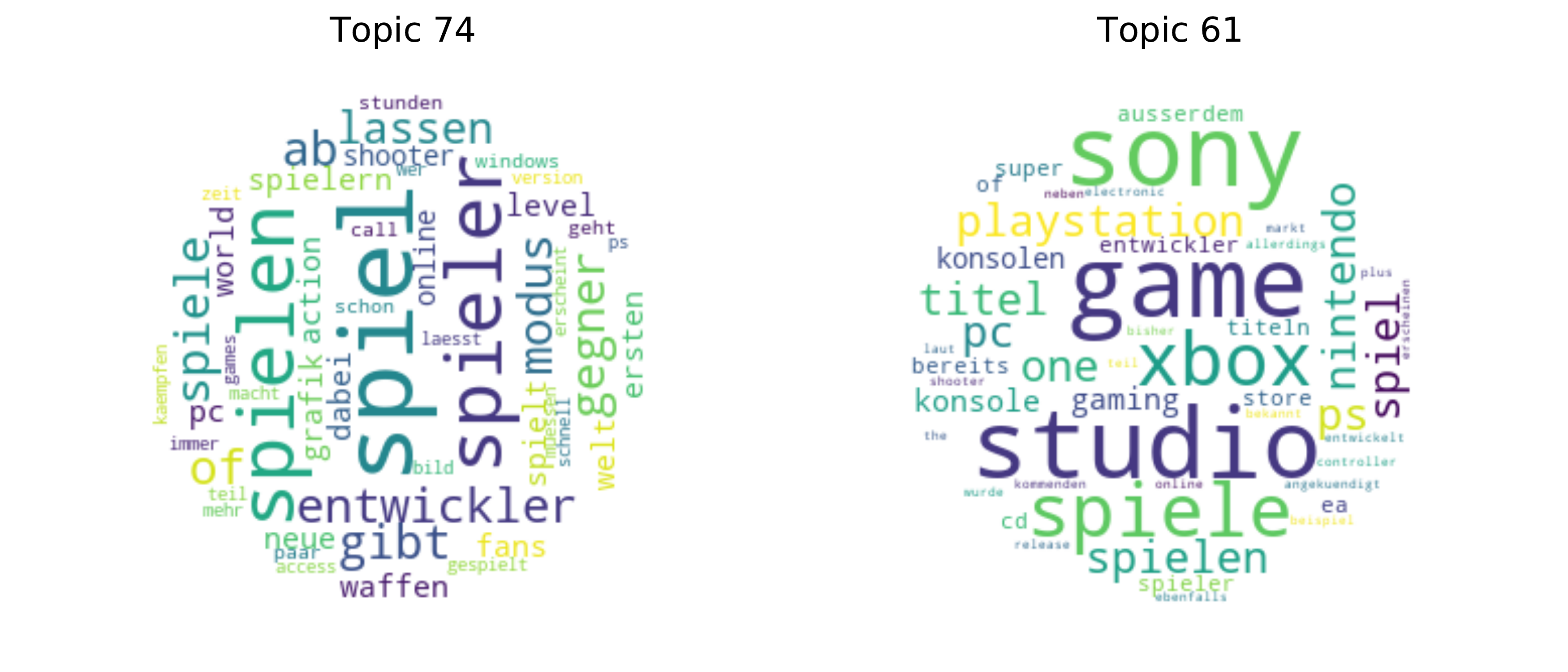}
        \caption{Similar topics in DGP 3}
        \label{fig:similar_topics_dgp3}
    \end{figure}
\end{enumerate}

\section{Recall and Precision }\label{Appendix_RecallPrecision}
\begin{figure}[H]
\centering
\includegraphics[width=0.95\textwidth]{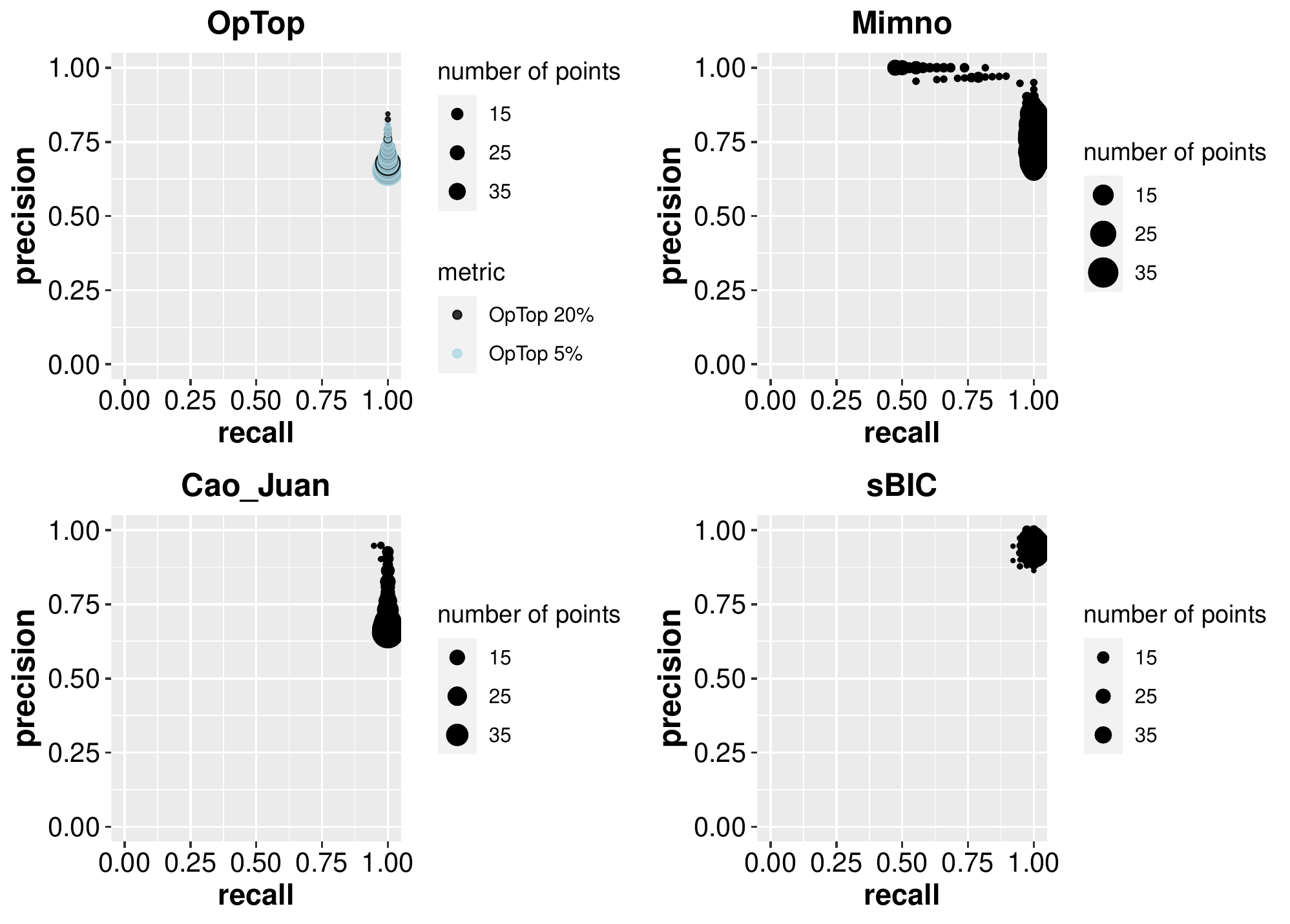}
\caption{Precision and recall for DGP1}
\label{fig: pre_rec_dgp1}
\end{figure}

\begin{figure}[H]
\centering
\includegraphics[width=0.95\textwidth]{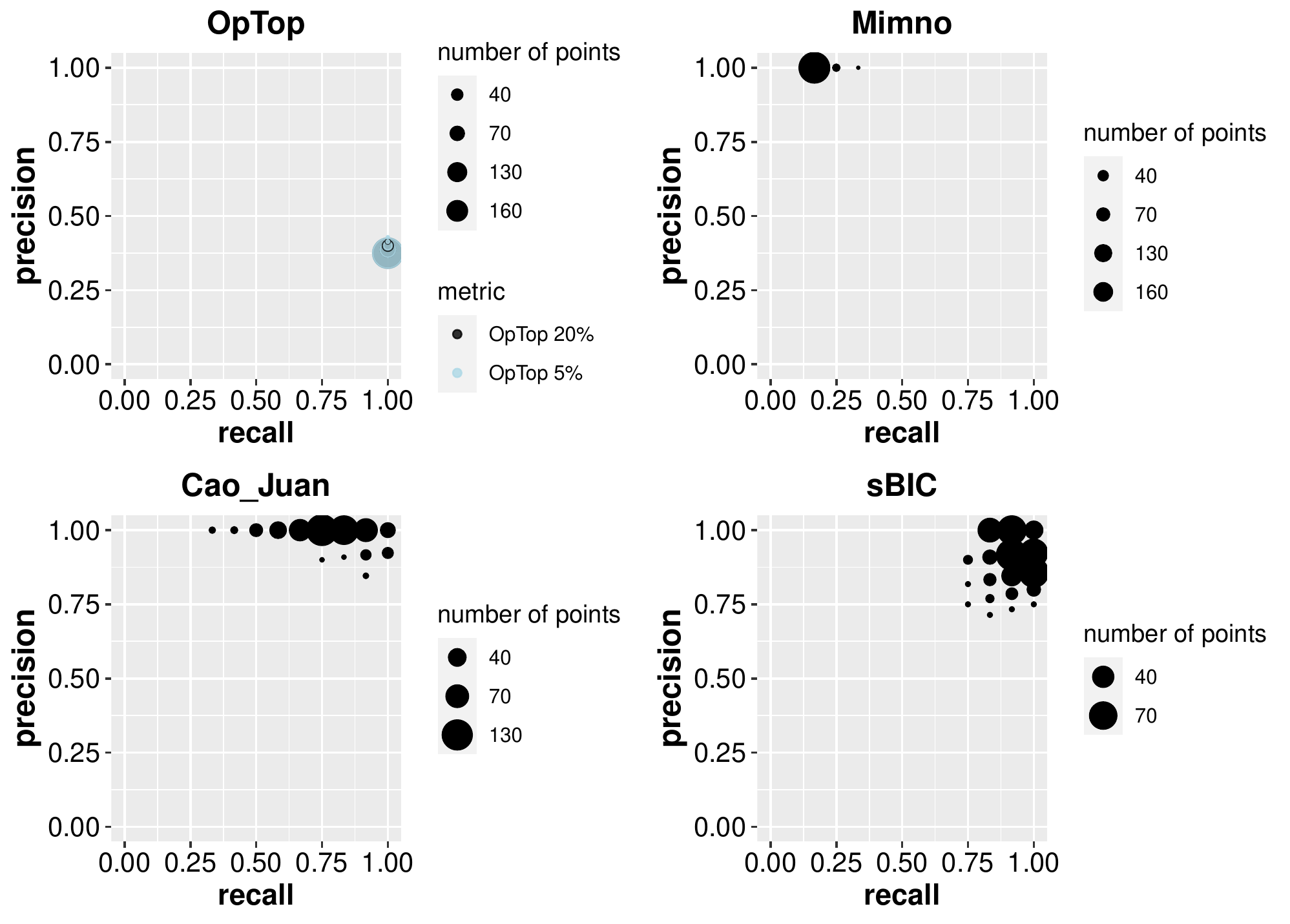}
\caption{Precision and recall for DGP2}
\label{fig: pre_rec_dgp2}
\end{figure}

\begin{figure}[H]
\centering
\includegraphics[width=0.95\textwidth]{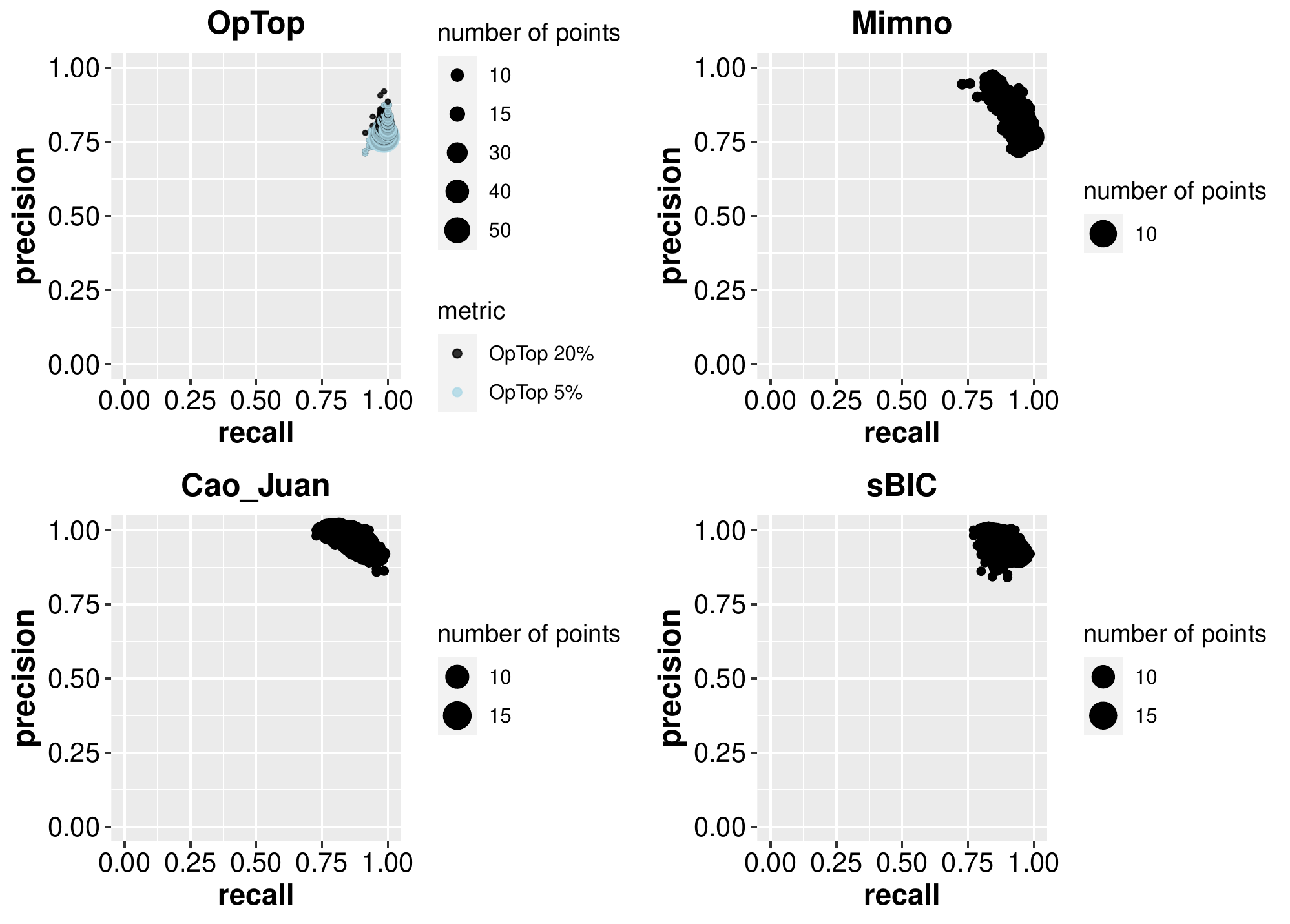}
\caption{Precision and recall for DGP3}
\label{fig: pre_rec_dgp3}
\end{figure}

Figures~\ref{fig: pre_rec_dgp1}, \ref{fig: pre_rec_dgp2}, and~\ref{fig: pre_rec_dgp3} exhibit the scatter plots of recall and precision values for each DGP separately. Thereby, each point corresponds to one of the simulated corpora. Consequently, there is a total of 300 points in each plot. However, the evaluation metrics considered may result in the same recall and precision scores for multiple corpora. Thus, some points may overlap.

\clearpage

\section{Weighted Recall \& Precision}\label{Appendix_WeightedRecall}

As a robustness analysis, we report results for alternative definitions of recall and precision. We identified a true positive (TP) for the measures in Section~\ref{sec:topic_content}, when the similarity of matched topics was above a predefined threshold. Here, we use the actual cosine similarity scores instead, which would be close to 1 for good matches. Hence, recall and precision values are calculated as follows:

\begin{equation}
    \text{Recall}=\frac{\sum_{i=1}^{n}\text{cosine\_similarity\_score}_{i}}{K_{true}},
\end{equation}

\begin{equation}
    \text{Precision}=\frac{\sum_{i=1}^{n}\text{cosine\_similarity\_score}_{i}}{K_{metric}},
\end{equation}

\noindent where $K_{true}$ is the true number of topics in a particular DGP. $K_{metric}$ is the proposed number of topics for the evaluation metric considered. The numerator contains the sum of cosine similarity values of all the $n$ identified matches. Therefore, recall presents the average cosine similarity value among the matches relative to the true number of topics. Precision presents the average cosine similarity value between the matches relative to the estimated number of topics.

Table~\ref{tab:weighted_recall_precision} summarizes the recall, precision, and F1 score values for this alternative definitions of recall and precision. As expected, the values are smaller than the values shown in Table~\ref{tab: recall_best_match} for the original definitions, but the qualitative findings about the relative performance of the different criteria remain unchanged. According to the F1 scores, sBIC performs best for DGP 1 and DGP 2, while the average F1 scores are quite similar for all the considered metrics in DGP 3, still with a minor advantage for Cao\_Juan and sBIC.

\begin{table}[H]
    \centering
    \begin{tabular}{|l|l|r|r|r|r|r|r|}
\toprule \hline
  &      & \multicolumn{2}{l|}{Recall} & \multicolumn{2}{l|}{Precision} & \multicolumn{2}{l|}{F1} \\ \cline{1-8}
 data   &   metric    &                   mean &   std &                      mean &   std &               mean &   std \\ \cline{1-8}
\midrule
\multirow{5}{*}{DGP1} & Cao\_Juan &                   0.99 &  0.00 &                      0.71 &  0.07 &               0.82 &  0.04 \\ \cline{2-8}
     & Mimno &                   0.95 &  0.13 &                      0.76 &  0.08 &               0.83 &  0.07 \\ \cline{2-8}
     & OpTop 20\% &                   0.98 &  0.00 &                      0.67 &  0.03 &               0.80 &  0.02 \\ \cline{2-8}
     & OpTop 5\% &                   0.98 &  0.00 &                      0.67 &  0.03 &               0.80 &  0.02 \\ \cline{2-8}
     & sBIC &                   0.99 &  0.02 &                      0.93 &  0.03 &               0.96 &  0.02 \\ \cline{2-8}
\cline{1-8}
\multirow{5}{*}{DGP2} & Cao\_Juan &                   0.76 &  0.14 &                      0.96 &  0.03 &               0.84 &  0.10 \\ \cline{2-8}
     & Mimno &                   0.11 &  0.02 &                      0.66 &  0.02 &               0.19 &  0.02 \\ \cline{2-8}
     & OpTop 20\% &                   1.00 &  0.00 &                      0.38 &  0.01 &               0.55 &  0.01 \\ \cline{2-8}
     & OpTop 5\% &                   1.00 &  0.00 &                      0.38 &  0.01 &               0.55 &  0.01 \\ \cline{2-8}
     & sBIC &                   0.92 &  0.07 &                      0.91 &  0.06 &               0.91 &  0.05 \\ \cline{2-8}
\cline{1-8}
\multirow{5}{*}{DGP3} & Cao\_Juan &                   0.85 &  0.06 &                      0.94 &  0.02 &               0.89 &  0.03 \\ \cline{2-8}
     & Mimno &                   0.92 &  0.04 &                      0.81 &  0.05 &               0.86 &  0.02 \\ \cline{2-8}
     & OpTop 20\% &                   0.98 &  0.02 &                      0.78 &  0.03 &               0.87 &  0.02 \\ \cline{2-8}
     & OpTop 5\% &                   0.98 &  0.02 &                      0.78 &  0.02 &               0.87 &  0.02 \\ \cline{2-8}
     & sBIC &                   0.86 &  0.05 &                      0.92 &  0.03 &               0.89 &  0.03 \\ \cline{1-8}
\bottomrule
\end{tabular}

    \caption{Descriptive statistics of recall, precision, and F1 scores based on cosine similarity}
    \label{tab:weighted_recall_precision}
\end{table}

\begin{figure}[H]
\centering
\includegraphics[width=0.8\textwidth]{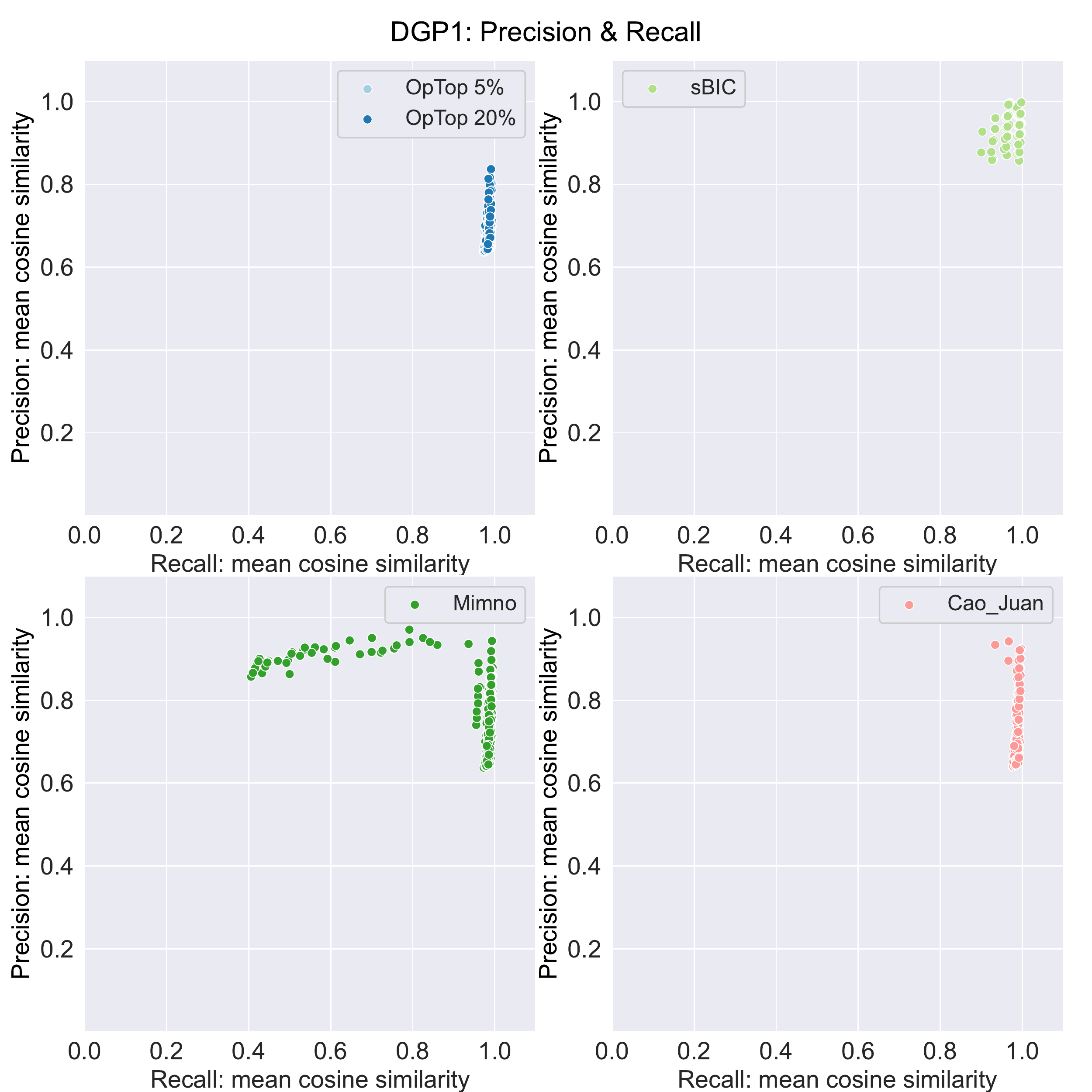}
\caption{Precision and recall based on cosine similarity for DGP1}
\label{fig: weighted_dgp1}
\end{figure}

\begin{figure}[H]
\centering
\includegraphics[width=0.8\textwidth]{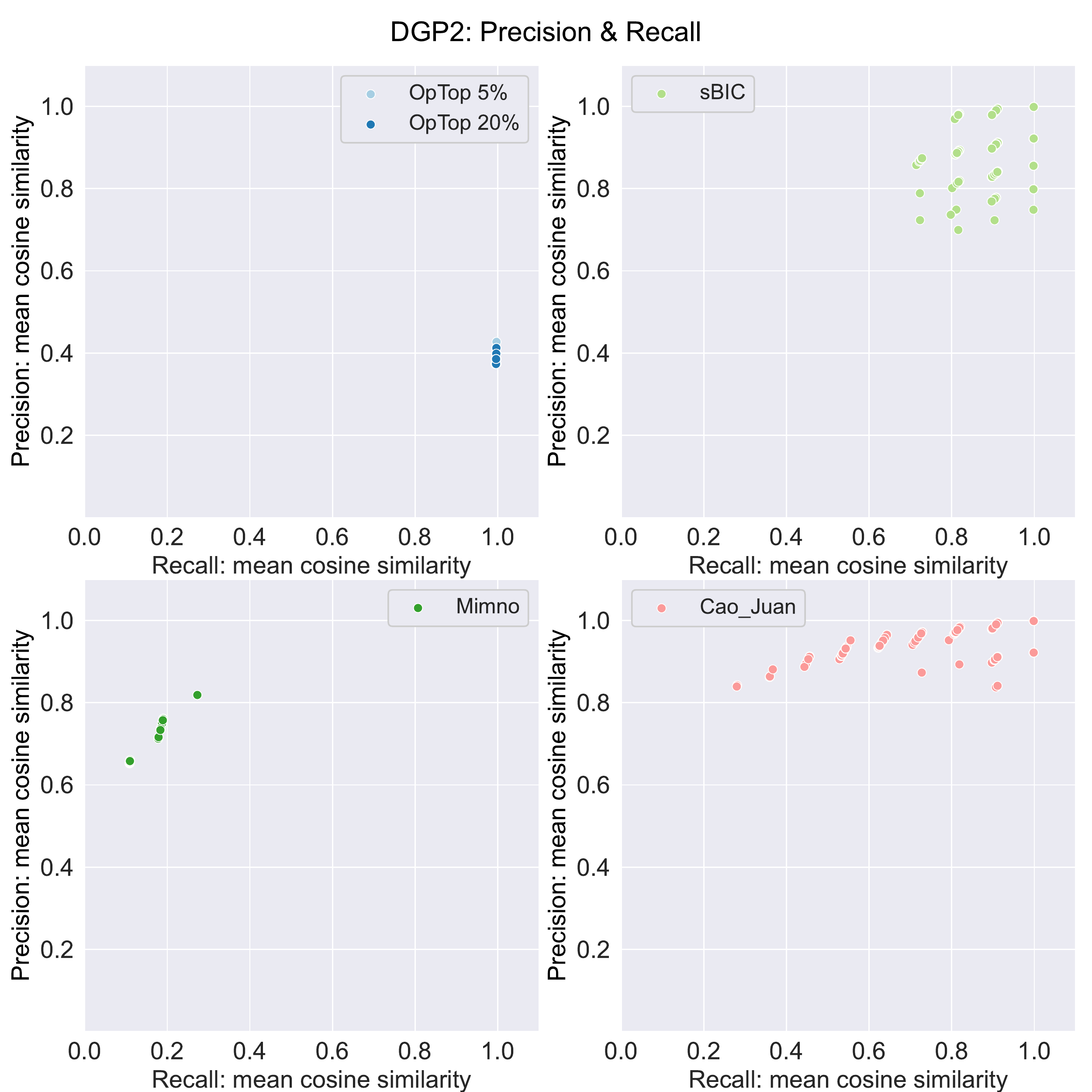}
\caption{Precision and recall based on cosine similarity for DGP2}
\label{fig: weighted_dgp2}
\end{figure}

\begin{figure}[H]
\centering
\includegraphics[width=0.8\textwidth]{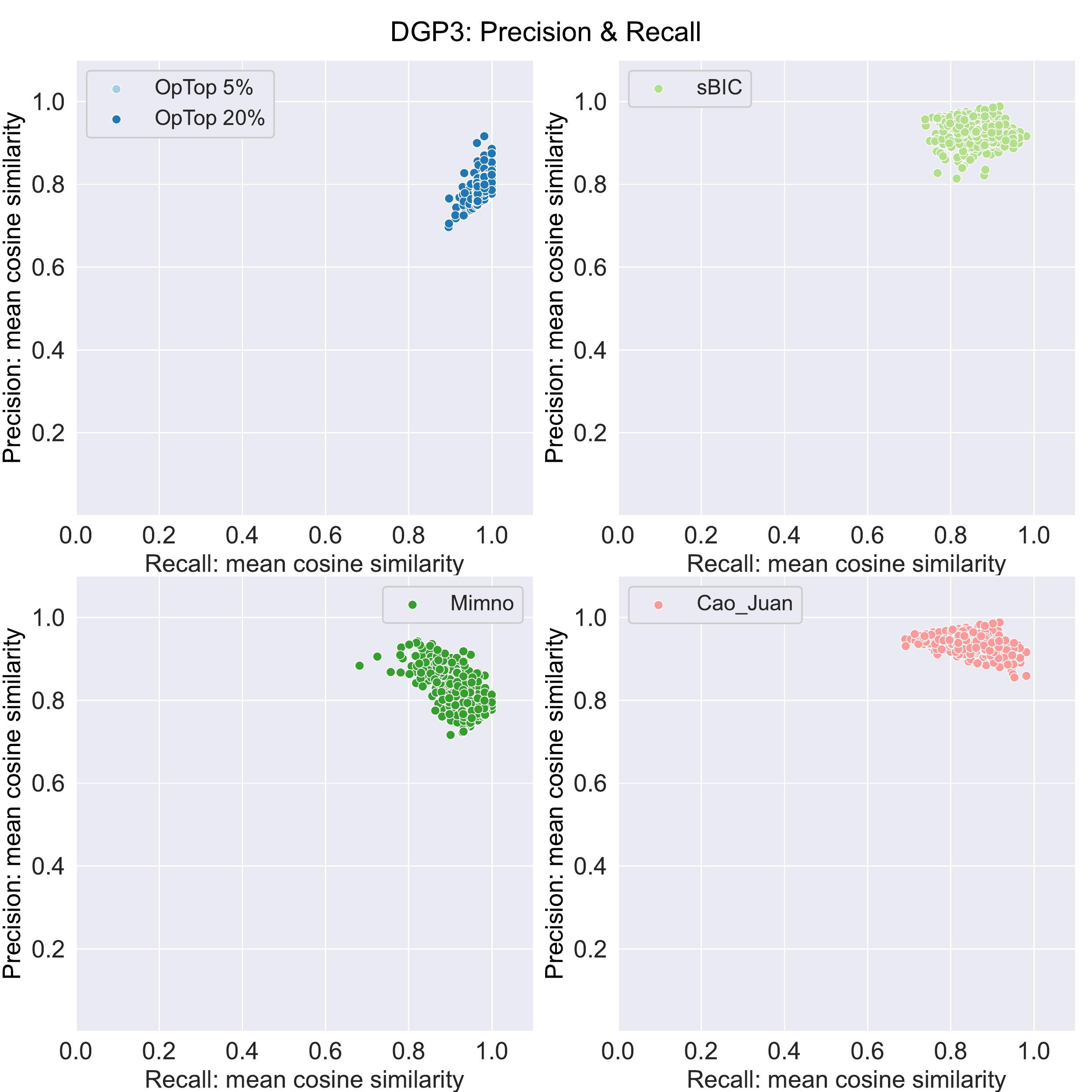}
\caption{Precision and recall based on cosine similarity for DGP3}
\label{fig: weighted_dgp3}
\end{figure}

While recall and precision values of our standard implementation are discrete leading to clustering of points in the scatter plots shown in Appendix~\ref{Appendix_RecallPrecision}, the weighted recall and precision values reported in this section are continuous and each point is actually unique due to the differences in the cosine values, although these might be minor. Therefore, we do not use the type of plots from Appendix~\ref{Appendix_RecallPrecision} taking into account the clustering, but standard scatter plots in Figures~\ref{fig: weighted_dgp1}, \ref{fig: weighted_dgp2}, and \ref{fig: weighted_dgp3}.

\end{appendices}

\end{document}